# ConcertoRL: An Innovative Time-Interleaved Reinforcement Learning Approach for Enhanced Control in Direct-Drive Tandem-Wing Vehicles

Minghao Zhang[1], Bifeng Song[2,*], Changhao Chen[3], Xinyu Lang[4]

[1] School of Aeronautics, Northwestern Polytechnical University, Xi'an 710072, China;

[2*] School of Aeronautics, Northwestern Polytechnical University, Xi'an 710072, China;

[3] School of Aeronautics, Northwestern Polytechnical University, Xi'an 710072, China;

[4] School of Aeronautics, Northwestern Polytechnical University, Xi'an 710072, China;

*Corresponding author(s). E-mail(s): sbf@nwpu.edu.cn (Bifeng Song);

Contributing authors:

zhangminghao2@mail.nwpu.edu.cn;

chenchanghao@mail.nwpu.edu.cn;

LANGXINYU@mail.nwpu.edu.cn

## Abstract

In control problems for insect-scale direct-drive experimental platforms under tandem wing influence, the precision and safety of control during plug-and-play online training and control processes are paramount.

The primary challenge facing existing reinforcement learning models is their limited safety in the exploration process and the stability of the continuous training process.

Addressing these challenges, we introduce the ConcertoRL algorithm to enhance control precision and stabilize the online training process, which consists of two main innovations: a time-interleaved mechanism to interweave classical controllers with reinforcement learning-based controllers aiming to improve control precision in the initial stages, a policy composer organizes the experience gained from previous learning to ensure the stability of the online training process.

This paper conducts a series of experiments. First, experiments incorporating the time-interleaved mechanism demonstrate a substantial performance boost of approximately 70% over scenarios without reinforcement learning enhancements and a 50% increase in efficiency compared to reference controllers with doubled control frequencies. These results highlight the algorithm's ability to create a synergistic effect that exceeds the sum of its parts. Second, Ablation studies on the policy composer further reveal that this module significantly enhances the stability of ConcertoRL during online training. Lastly, experiments on the universality of the current ConcertoRL algorithm framework demonstrate its compatibility with various classical controllers, consistently achieving excellent control outcomes.

ConcertoRL sets a new benchmark in control effectiveness for challenges posed by direct-drive platforms under tandem wing influence and establishes a comprehensive framework for integrating classical and reinforcement learning-based control methodologies.

# 1 Introduction

As critical supporting technologies advance, Hover-capable flapping wing aircraft's advantages in size and efficiency over equivalent rotary-wing aircraft have become increasingly apparent[1-4]. Consequently, researchers have developed a large number of flapping-wing aircraft such as DelFly Nimble[5], Quad-thopter[6], KUBeetle[7], and Nano-Hummingbird[8],. In this context, our team has developed the DDD-1 aircraft[9], as shown in Fig. 1, which faces significant challenges due to the unsteady aerodynamic interference caused by its tandem wings. This interference complicates the coordinated movement of the wings and adversely affects the aerodynamic performance, leading to a noticeable decline in flight capabilities. So, conducting a detailed aerodynamic study on a dedicated experimental platform is imperative to improve its design and performance.

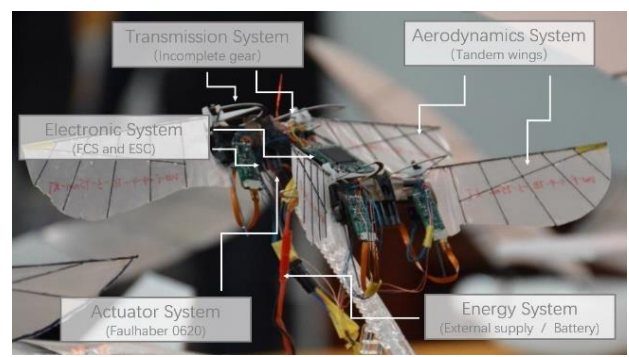

**Fig. 1.** Illustration of system integration of the DDD-1

As shown in Fig. 2, Aerodynamic experimental platforms play a pivotal role in the study of aerodynamic systems, with a wealth of research already established in this domain. For instance, James R. Usherwood et al.[10] developed a semi-direct drive experiment platform installed in a mineral oil tank, featuring sensors mounted at the wing root and motion transmitted via a gearbox. This platform facilitated the investigation of the impact of phase differences between tandem wings on lift and power consumption characteristics during hover. Similarly, Hiroto Nagai et al.[11] constructed a direct drive experimental platform placed in a water tunnel, with a torque sensor installed on the output shaft. Their research focused on the effects of different phases on the overall lift, required power consumption, and pitching moment under tandem wing conditions.

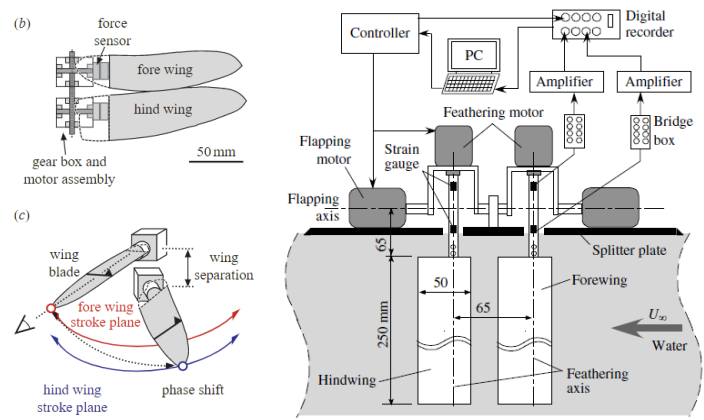

**Fig. 2.** Typical direct drive test bench aerodynamic experimental platforms

While existing experimental platforms, through direct or semi-direct drive mechanisms, ensure the diversity of researchable trajectories and enable the separate measurement of wing operational characteristics, they predominantly conduct experiments in viscous fluids like mineral oil under Reynolds number similarity conditions using rigid materials for wings. This approach, aimed at reducing the complexity of experimental design and measurement by lowering flapping frequencies and minimizing the effects of wing deformation, limits the direct applicability of their findings to the design of wings and trajectories for actual flight vehicles. In response to these limitations, specifically for the DDD-1 aircraft, this study introduces an experiment platform designed for high-frequency flapping in air. This platform introduces the challenges of high-frequency control in nonlinear, unsteady systems.

Numerous researchers have ventured into the domain of controlling nonlinear, unsteady systems[12], broadly dividing the existing methodologies into classical and artificial intelligence control algorithms.

Classical control strategies encompass robust control[13], model predictive control[14], and optimal control[15]. Among these, model-free, data-driven control algorithms, epitomized by PID, have found widespread application across various control challenges[16-18]. While these methodologies have achieved notable success in specific control scenarios, a pressing issue in the context of experiment platforms for dynamic systems is the inherent limitations they present: initially, there is an absence of models to base the control on, and throughout the online training phase, there is a critical need to prevent damage to the apparatus and avoid frequent stops and starts, which could reduce the experimental platform's lifespan. These constraints pose significant challenges in tuning the parameters of classical controllers through experimental means.

The RL offers a versatile framework within artificial intelligence control algorithms, presenting a data-driven approach to learning behaviors through interaction[19]. This method has demonstrated commendable control performance across various problems[20]. However, the practical deployment of RL in robotic and control systems is far from straightforward, as highlighted by existing literature[21]. Key challenges emanate from the efficiency of its training processes and the safety of its exploration strategies[21, 22].

One strategy to address the challenges of training efficiency and exploration safety in the RL involves online exploration entirely, opting to learn from historical data without further system interaction, a method known as offline reinforcement learning[23]. Yang et al.[24] have pioneered a model-based reinforcement learning framework designed explicitly for legged robot locomotion. This innovative framework, emphasizing safety and high efficiency, achieves robust walking capabilities utilizing just 4.5 minutes of training data. It integrates prior knowledge of leg trajectories and employs a novel loss function tailored for long-term dynamics modeling. This approach markedly improves the sample efficiency rates of traditional model-free methods and facilitates versatile task adaptation via minor adjustments to the reward function. Despite these advancements, the framework's performance is inherently contingent upon the quality of the static dataset used for offline learning, with significant challenges including coverage issues, distribution shifts, and constraints on policy improvement.

Strategies involving the adjustment of rewards or the introduction of constraints to limit the exploration process have been explored to address the challenges of training efficiency and exploration safety from another perspective. Achiam et al.[25]. introduce

Constrained Policy Optimization (CPO), a groundbreaking method that ensures safety and efficiency in the training process. By integrating constraints directly with the reward functions, CPO enables the formulation of neural network policies tailored for high-dimensional control tasks. This approach guarantees near-constraint satisfaction with robust theoretical support and enhances safety without detracting from performance efficiency. However, a notable limitation is its tendency to prioritize adherence to constraints over minimizing costs during the training process.

Guided Policy Search (GPS) presents a solution by learning control policies directly from complex, high-dimensional environments to enhance training efficiency and exploration safety. It employs trajectory-centric optimization techniques and reformulates policy search into supervised learning, thus ensuring efficient usage of samples and providing guarantees for local convergence. Zhang et al.[26] further advance this domain by integrating Model Predictive Control (MPC) with reinforcement learning within the GPS framework. This integration facilitates training deep neural network policies utilizing data generated by MPC under full-state observation scenarios. This methodology enables effective robot control based on raw sensor data, markedly reducing the computational burden. However, a limitation of guided policy search algorithms, including this approach, is their ultimate convergence to the original 'reference policy,' with performance constrained by the reference controller.

To address the critical challenges of training efficiency, exploration safety, and training stability, Sergey Levine and colleagues[27] introduced the APRL framework, which capitalizes on deep reinforcement learning to swiftly teach a quadruped robot to walk in real-world conditions while continually enhancing its capabilities. This framework marks a significant stride towards the autonomous acquisition of complex behaviors by robots in practical scenarios, promoting efficiency and safety in the learning process. Despite acknowledging that most random actions could lead to catastrophic failures, the APRL framework introduces a dynamic curriculum mechanism based on familiarity measures to mitigate these risks. However, the framework still faces challenges in achieving immediate superior control upon deployment, with exploration risks remaining and direct plug-and-play control still needing to realize optimal performance outcomes.

In response to the challenges associated with plug-and-play online training for direct-drive experimental platforms influenced by tandem wing dynamics, the application of Reinforcement Learning to the motion control problems of such platforms is explored in this study. A comprehensive scheme is provided, and notable results are achieved. The main contributions of our work are outlined as follows:

1. To the best of our knowledge, this is the inaugural exploration of plug-and-play online training for controlling systems subjected to the strong non-linearity and unsteadiness introduced by tandem wing interference. This exploration is particularly significant given the control challenges posed at the insect scale with tandem wing disturbances, which are substantially more complex than those encountered with conventional dual-wing flapping vehicles, quadrotors, and fixed wings.
2. A dynamic system model for direct-drive experimental platforms under tandem wing influence has been constructed, encapsulating the primary characteristics of these control issues by integrating multi-body dynamics and unsteady load considerations.
3. A novel reinforcement learning algorithm, ConcertoRL, has been designed to refine control precision and enhance the stability of the online training process. This approach has demonstrated a significant improvement in performance within the initial 500 steps of interaction compared to the SAC algorithm, showcasing enhanced sample efficiency and safety. The potential for fully implementing RL controllers for real-world experimental platforms as plug-and-play control has been significantly advanced.
4. To augment control precision in the initial stages, a time-interleaved mechanism that alternates classical controllers with reinforcement learning-based controllers has been devised and validated through ablation experiments for its effectiveness.
5. The stability of the online training process has been fortified through the design of a policy composer, which effectively organizes the experience gained from previous learning, as confirmed by ablation studies.
6. Comprehensive adaptability studies have been conducted to ascertain the compatibility of the current algorithm with multiple states and various reference controllers under different parameters.

The remainder of this article is structured as follows. Section II provides a detailed description of the problem being addressed. Section III introduces the ConcertoRL algorithm, detailing its methodology and theoretical underpinnings. Section IV presents the experiments conducted to evaluate the effectiveness of the ConcertoRL algorithm, including the experimental platform, data analysis, and discussion of results. Finally, Section V concludes the article with a summary of the findings and contributions.

## 2 Problem description

### 2.1 Control problem description

This study addresses the challenging problem of tracking control under conditions of nonlinearity and unsteady with partially observable states, specifically focusing on the trajectory tracking of four wings on a direct-drive testbed. The inherent challenge lies in the direct-drive mechanism's inability to utilize transmission mechanisms for generating a consistent motion trajectory, necessitating an effective strategy to characterize the wings' motion trajectory accurately.

To tackle this issue, we employ a motion strategy organized around the concept of Central Pattern Generators (CPGs)[28] to generate the desired trajectory. The following equation describes the expected motion of each wing.

$$\phi_{EXP-n-i} = A_i \cdot sin(2\pi f \cdot t + \varphi_i) \quad (1)$$

where $\phi_{EXP-n-i}$ denotes the expected flapping angle position of the i-th wing for the next n steps, $A_i$ denotes the desired flapping amplitude for the i-th wing, $f$ represents the flapping frequency, and $\varphi_i$ signifies the phase difference of the i-th wing.

### 2.2 Modeling direct-drive platform under tandem wing influence

Current modeling efforts are focused on addressing the challenges posed by the nonlinear and unsteady aerodynamic characteristics associated with tandem wing interference and the difficulties in control due to partially observable states caused by the inability to observe critical rotation angles directly.

As shown in Fig. 3, the experimental platform is divided into 13 key components: one bench, four motors, four springs, and four wings. The rig is rigidly attached to the ground at the $O_G^R$ point; the four motors and springs are interconnected and securely mounted on the bench; and the four wings are individually connected to the rig at the $O_{w,1}^R$, $O_{w,2}^R$, $O_{w,3}^R$, and $O_{w,14}^R$ points, respectively, through constraints with the motors. In the construction of the aforementioned system model, due to the experimental rig's symmetry along the $X_G Z_G$ plane, the points $O_{w,1}$ and $O_{w,2}$ are symmetrical relative to the $X_G Z_G$ plane, as are $O_{w,4}$ and $O_{w,3}$.

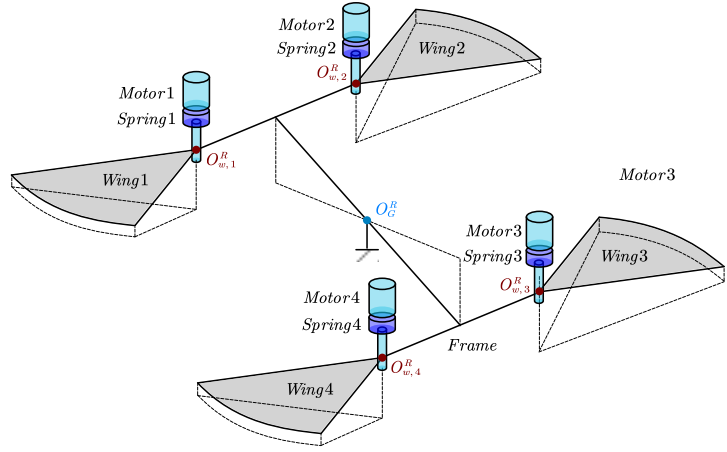

**Fig. 3.** Schematic diagram of the main components of the experimental platform

Based on the analysis, the research subject encompasses eight degrees of freedom: $\phi_{w,1}$、$\theta_{w,1}$、$\phi_{w,2}$、$\theta_{w,2}$、$\phi_{w,3}$、$\theta_{w,3}$、$\phi_{w,4}$、$\theta_{w,4}$. This presents a complex multi-rigid body challenge, necessitating applying the Lagrangian method to construct the system's dynamic equations[29]. Consequently, the simulation system framework is established in Fig. 4.

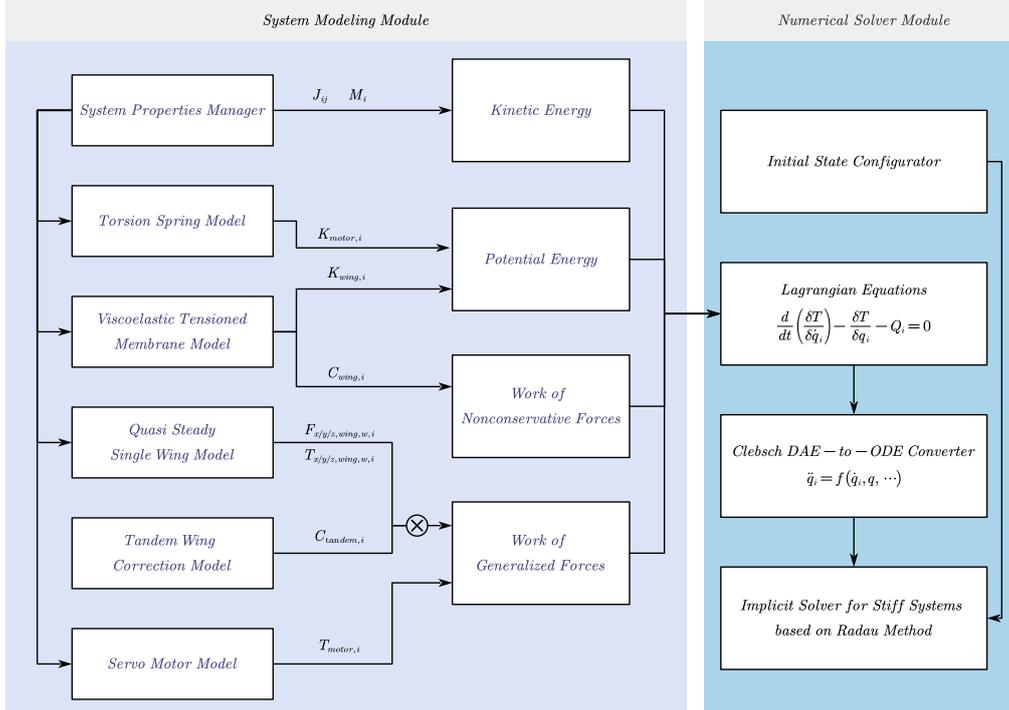

**Fig. 4.** Simulation system framework

Given the inherent complexity of the system dynamics equations derived from the Lagrangian equation method, this study simplifies the process by substituting specific values for the constant variables presented in Appendix A. Consequently, the dynamics equations for the eight degrees of freedom system are as follows:

$$\ddot{\phi}_{w,1} = 3241491.09 \cdot C_{wing,1} \cdot \dot{\theta}_1 - 1620745.54 \cdot K_{motor,1} \cdot \phi_1 \\
+ 6482982.17 \cdot K_{wing,1} \cdot \theta_1 + 1620745.54 \cdot T_{motor,1} \\
- 3241491.09 \cdot T_{y,Twing,w,1} + 1620745.54 \\
\cdot T_{z,Twing,w,1} \quad (2)$$

$$\ddot{\phi}_{w,2} = 3241491.09 \cdot C_{wing,2} \cdot \dot{\theta}_2 - 1620745.54 \cdot K_{motor,2} \cdot \phi_2 \\
+ 6482982.17 \cdot K_{wing,2} \cdot \theta_2 + 1620745.54 \cdot T_{motor,2} \\
- 3241491.09 \cdot T_{y,Twing,w,2} + 1620745.54 \\
\cdot T_{z,Twing,w,2} - 5091722.29 \cdot K_{motor,2} \quad (3)$$

$$\ddot{\phi}_{w,3} = 3241491.09 \cdot C_{wing,2} \cdot \dot{\theta}_2 - 1620745.54 \cdot K_{motor,3} \cdot \phi_3 \\
+ 6482982.17 \cdot K_{wing,3} \cdot \theta_3 + 1620745.54 \cdot T_{motor,3} \\
- 3241491.09 \cdot T_{y,Twing,w,3} + 1620745.54 \\
\cdot T_{z,Twing,w,3} - 5091722.29 \cdot K_{motor,3} \quad (4)$$

$$\ddot{\phi}_{w,4} = 3241491.09 \cdot C_{wing,4} \cdot \dot{\theta}_4 - 1620745.54 \cdot K_{motor,4} \cdot \phi_4 \\
+ 6482982.17 \cdot K_{wing,4} \cdot \theta_4 + 1620745.54 \cdot T_{motor,4} \\
- 3241491.09 \cdot T_{y,Twing,w,4} + 1620745.54 \\
\cdot T_{z,Twing,w,4} \quad (5)$$

$$\ddot{\theta}_{w,1} = -39816315.51 \cdot C_{wing,1} \cdot \dot{\theta}_1 + 3241491.09 \cdot K_{motor,1} \cdot \phi_1 \\
- 79632631.01 \cdot K_{wing,1} \cdot \theta_1 - 3241491.09 \\
\cdot T_{motor,1} + 39816315.51 \cdot T_{y,Twing,w,1} \\
- 3241491.09 \cdot T_{z,Twing,w,1} \quad (6)$$

$$\ddot{\theta}_{w,2} = -39816315.51 \cdot C_{wing,2} \cdot \dot{\theta}_2 + 3241491.09 \cdot K_{motor,2} \cdot \phi_2 \\
- 79632631.01 \cdot K_{wing,2} \cdot \theta_2 - 3241491.09 \\
\cdot T_{motor,2} + 39816315.51 \cdot T_{y,Twing,w,2} \\
- 3241491.09 \cdot T_{z,Twing,w,2} + 10183444.58 \\
\cdot K_{motor,2} \quad (7)$$

$$\ddot{\theta}_{w,3} = -39816315.51 \cdot C_{wing,3} \cdot \dot{\theta}_3 + 3241491.09 \cdot K_{motor,3} \cdot \phi_3 \\
- 79632631.01 \cdot K_{wing,3} \cdot \theta_3 - 3241491.09 \\
\cdot T_{motor,3} + 39816315.51 \cdot T_{y,Twing,w,3} \\
- 3241491.09 \cdot T_{z,Twing,w,3} + 10183444.58 \\
\cdot K_{motor,3} \quad (8)$$

$$\ddot{\theta}_{w,4} = -39816315.51 \cdot C_{wing,4} \cdot \dot{\theta}_4 + 3241491.09 \cdot K_{motor,4} \cdot \phi_4 \\
- 79632631.01 \cdot K_{wing,4} \cdot \theta_4 - 3241491.09 \\
\cdot T_{motor,4} + 39816315.51 \cdot T_{y,Twing,w,4} \\
- 3241491.09 \cdot T_{z,Twing,w,4} \quad (9)$$

$$\ddot{\theta}_{w,4} = -39816315.51 \cdot C_{wing,4} \cdot \dot{\theta}_4 + 3241491.09 \cdot K_{motor,4} \cdot \phi_4$$
$$- 79632631.01 \cdot K_{wing,4} \cdot \theta_4 - 3241491.09$$
$$\cdot T_{motor,4} + 39816315.51 \cdot T_{y,\text{Twing},w,4} \quad (10)$$
$$- 3241491.09 \cdot T_{z,\text{Twing},w,4}$$

$$\begin{bmatrix} 0 \\ T_{y,\text{Twing},w,1} \\ T_{z,\text{Twing},w,1} \end{bmatrix} = \begin{bmatrix} 0 \\ T_{y,\text{wing},w,1} \\ T_{z,\text{wing},w,1} \end{bmatrix} \cdot (C_{tandem,1} + 1) \quad (11)$$

$$\begin{bmatrix} 0 \\ T_{y,\text{Twing},w,2} \\ T_{z,\text{Twing},w,2} \end{bmatrix} = \begin{bmatrix} 0 \\ T_{y,\text{wing},w,2} \\ T_{z,\text{wing},w,2} \end{bmatrix} \cdot (C_{tandem,2} + 1) \quad (12)$$

$$\begin{bmatrix} 0 \\ T_{y,\text{Twing},w,3} \\ T_{z,\text{Twing},w,3} \end{bmatrix} = \begin{bmatrix} 0 \\ T_{y,\text{wing},w,3} \\ T_{z,\text{wing},w,3} \end{bmatrix} \cdot (C_{tandem,3} + 1) \quad (13)$$

$$\begin{bmatrix} 0 \\ T_{y,\text{Twing},w,4} \\ T_{z,\text{Twing},w,4} \end{bmatrix} = \begin{bmatrix} 0 \\ T_{y,\text{wing},w,4} \\ T_{z,\text{wing},w,4} \end{bmatrix} \cdot (C_{tandem,4} + 1) \quad (14)$$

where $C_{wing,i}$ represents the damping generated during the tensioning process of the membrane part of the $i-$th wing. $K_{wing,i}$ denotes the stiffness resulting from the tensioning process of the membrane part of the $i-$th wing. $K_{motor,i}$ is the stiffness coefficient of the spring installed at the servo motor of the $i-$th wing. $T_{motor,i}$ specifies the output torque of the $i-$th servo motor. $C_{tandem,i}$ indicates the influence coefficient of the interference from tandem wings on the $i-$th wing. $T_{y,\text{wing},w,i}$ is the torque around the $Y_{w,i}$ axis experienced by the wing surface, generated based on the aerodynamic force of the single wing. $T_{z,\text{wing},w,i}$ is the torque around the $Z_{w,i}$ axis experienced by the wing surface, generated based on the aerodynamic force of the single wing.

## 2.3 Aerodynamic modeling of single wings

This section focuses on determining the parameters $T_{y,\text{wing},w,i}$ and $T_{z,\text{wing},w,i}$ within the framework of the proposed model.

The study employs a flapping wing configuration as depicted in Figure X, wherein the complete flapping wing is bifurcated into two distinct components: a rigid plate, which executes flapping and rotation movements to generate aerodynamic forces, and a flexible membrane, which, though not contributing to aerodynamic force generation, imparts restraining forces upon being tensioned.

This research aims to construct a quasi-steady aerodynamic force model for a single wing, utilizing the quasi-steady aerodynamic estimation model introduced by

Ellington and colleagues[30], complemented by the empirical findings from Lee and colleagues [31]. The aerodynamic model primarily encompasses translational aerodynamic forces and moments, rotational aerodynamic forces and moments, and added mass aerodynamic forces and moments.

In terms of geometric parameters, a trapezoidal outline, as shown in Fig. 5, has been selected for subsequent simulation endeavors. The flapping motion axis is aligned parallel to the wing root, whereas rotational motion occurs around the leading edge of the flapping wing. The distance from the flapping axis to the wingtip is denoted as $R$, with a wingspan of $B$, and a deviation between the wing root and the flapping axis of $\Delta R = R - B$. The chord lengths at the root and tip are specified as $c_R$ and $c_T$, respectively. For each wing segment $\Delta r$, the chord length as a function of the position $r$ relative to the wing root is defined as $c(r)$.

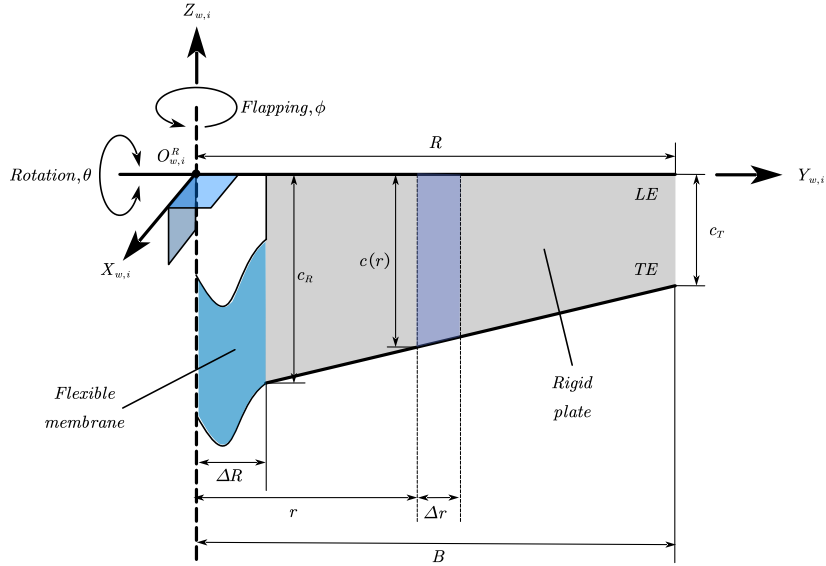

**Fig. 5.** Schematic representation of the geometric parameters and the definition of the wing surface coordinate system for a flapping wing, with the Leading Edge (LE) and the Trailing Edge (TE) clearly delineated.

In the modeling process, the mean chord length ($\bar{c}$), wing area ($S$), aspect ratio ($AR$), and the dimensionless second moment of area ($r_2$) are identified as critical morphological parameters. The definitions of these parameters are as follows:

$$\bar{c} = (c_R + c_T)/2 \tag{15}$$

$$S = \int_{\Delta R}^{R} c(r) dr \tag{16}$$

$$AR = \frac{R^2}{S} \tag{17}$$

$$r_2 = \frac{R_2}{R} = \frac{\sqrt{\frac{1}{S}\int_{\Delta R}^{R} c(r) \cdot r^2 \cdot dr}}{R} \tag{18}$$

In the computation of aerodynamic forces, the most critical parameters are the local angle of attack, $\alpha(r)$, and the local flow velocity, $U(r)$, for each wing segment. These can be calculated using the following formula:

$$\alpha(r) = actan(v_x(r)/v_z(r)) \tag{19}$$

$$U(r) = \sqrt{v_x(r)^2 + v_z(r)^2} \tag{20}$$

where $v_x(r)$ denotes the local velocity component along the x-axis of the wing surface coordinate system, $v_z(r)$ represents the local velocity component along the z-axis of the wing surface coordinate system.

Regarding the calculation of translational aerodynamic forces and moments, which serve as the principal source of aerodynamic forces during the flapping process of the flapping wing, the computational formula is as follows:

$$\Delta F_{x,tr,w,i} = \Delta F_{N,tr} = -1 \cdot C_{N,tr} \cdot 0.5 \cdot \rho \cdot U(r)^2 \cdot c(r) \cdot dr \tag{21}$$

$$\Delta F_{y,tr,w,i} = 0.0 \tag{22}$$

$$\Delta F_{z,tr,w,i} = \Delta F_{T,tr} = C_{T,tr} \cdot 0.5 \cdot \rho \cdot U(r)^2 \cdot c(r) \cdot dr \tag{23}$$

$$C_{N,tr} = [3.48 \cdot \sin(\alpha)] \tag{24}$$

$$C_{T,tr} = [0.4 \cdot \cos(2\alpha)^2] \tag{25}$$

where $\rho$ denotes the air density. The torque generated on each strip is:

$$\begin{bmatrix} \Delta T_{x,tr,i} \\ \Delta T_{y,tr,i} \\ \Delta T_{z,tr,i} \end{bmatrix} = \begin{bmatrix} 0.0 \\ r_{y,tr,w,i} \\ r_{z,tr,w,i} \end{bmatrix} \times \begin{bmatrix} \Delta F_{x,tr,w,i} \\ \Delta F_{y,tr,w,i} \\ \Delta F_{z,tr,w,i} \end{bmatrix} = \begin{bmatrix} +\Delta F_{z,tr,w,i} \cdot L_{S,tr,cop} \\ -\Delta F_{x,tr,w,i} \cdot L_{C,tr,cop} \\ -\Delta F_{x,tr,w,i} \cdot L_{S,tr,cop} \end{bmatrix} \tag{26}$$

$$L_{S,tr,cop} = r \tag{27}$$

$$L_{C,tr,cop} = 0.388 \cdot c(r) \tag{28}$$

The total aerodynamic force and moment on the wing surface, resulting from translational motion within the wing coordinate system, are determined through the following calculation:

$$F_{x,tr,w,i} = \int_{\Delta R}^{R} \Delta F_{x,tr,w,i} \tag{29}$$

$$F_{y,tr,w,i} = 0.0 \tag{30}$$

$$F_{z,tr,w,i} = \int_{\Delta R}^{R} \Delta F_{z,tr,w,i} \tag{31}$$

$$T_{x,tr,i} = \int_{\Delta R}^{R} \Delta T_{x,tr,i} \tag{32}$$

$$T_{y,tr,i} = \int_{\Delta R}^{R} \Delta T_{y,tr,i} \tag{33}$$

$$T_{z,tr,i} = \int_{\Delta R}^{R} \Delta T_{z,tr,i} \tag{34}$$

In calculating rotational aerodynamic forces and moments, these are primarily induced by the significant rotational motion of the flapping wing, attributed to the rotational circulation around the wing. The computational formula is as follows:

$$\Delta F_{x,rot,w,i} = -1 \cdot f_\alpha \cdot f_r \cdot C_{rot,1} \cdot \rho \cdot \dot\beta_z \cdot \dot\beta_y \cdot r \cdot c(r) \cdot dr \tag{35}$$

$$\Delta F_{x,rot,w,i} = f_\alpha \cdot f_r \cdot C_{rot,1} \cdot \rho \cdot v_x(r) \cdot v_z(r) \cdot c(r) \cdot dr + 2.67 \cdot \rho \\ \cdot \dot\beta_y \cdot |\dot\beta_y| \cdot \left[\int_{LE}^{TE} r \cdot x|x|dx\right] \cdot dr \tag{36}$$

$$f_\alpha = \begin{cases} +1, -45° < \alpha < 45° \\ -1, 135° < \alpha < 225° \\ \sqrt{2}\cos(\alpha), otherwise \end{cases} \tag{37}$$

$$\Delta F_{y,rot,w,i} = 0.0 \tag{38}$$
$$\Delta F_{z,rot,w,i} = 0.0 \tag{39}$$
$$C_{rot,1} = 0.842 - 0.507 \cdot Re^{-0.158} \tag{40}$$
$$\dot\beta_z = \dot\phi \tag{41}$$
$$\dot\beta_y = \dot\theta \tag{42}$$

where $\dot\beta_z$ represents the projection of the current wing surface angular velocity on the $Z_{w,i}$ axis; $\dot\beta_y$ denotes the projection of the current wing surface angular velocity on the $Y_{w,i}$ axis; $Re$ is the Reynolds number. The aerodynamic force and moment generated due to rotation are:

$$F_{x,rot,w,i} = \int_{\Delta R}^{R} \Delta F_{x,rot,w,i} \tag{43}$$

$$\begin{bmatrix} T_{x,rot,i} \\ T_{y,rot,i} \\ T_{z,rot,i} \end{bmatrix} = \begin{bmatrix} 0.0 \\ r_{y,rot,w,i} \\ r_{z,rot,w,i} \end{bmatrix} \times \begin{bmatrix} F_{x,rot,w,i} \\ F_{y,rot,w,i} \\ F_{z,rot,w,i} \end{bmatrix} = \begin{bmatrix} 0.0 \\ -\Delta F_{x,rot,w,i} \cdot L_{C,rot,cop} \\ -\Delta F_{x,rot,w,i} \cdot L_{S,rot,cop} \end{bmatrix} \tag{44}$$

$$F_{y,rot,w,i} = 0.993 \cdot R_2 \tag{45}$$
$$F_{z,rot,w,i} = 0.398 \cdot \bar{c} \tag{46}$$
$$F_{x,rot,w,i} = F_{x,rot,w,i} \tag{47}$$
$$F_{y,rot,w,i} = 0.0 \tag{48}$$
$$F_{z,rot,w,i} = 0.0 \tag{49}$$
$$T_{x,rot,i} = 0.0 \tag{50}$$
$$T_{y,rot,i} = -\Delta F_{x,tr,w,i} \cdot L_{C,tr,cop} \tag{51}$$

$$T_{z,rot,i} = -\Delta F_{x,tr,w,i} \cdot L_{S,tr,cop} \tag{52}$$

In calculating added mass aerodynamic forces and moments, these are predominantly induced by the interaction between the flapping wing and the surrounding air during acceleration or deceleration phases. The computational framework is as follows:

$$\begin{aligned}\Delta F_{x,add,w,i} = &-1 \cdot f_{\lambda,\alpha} \cdot f_{AR,\alpha} \cdot f_{a,a} \cdot \frac{\rho\pi}{4} \cdot \ddot{\beta}_z \cdot \sin(|\theta|) \cdot c(r)^2 \cdot r \cdot dr \\ &+ \ddot{\beta}_y \cdot c(r)^2 \cdot \left[\frac{LE(r)+TE(r)}{2} - rot_x(r)\right] \cdot dr\end{aligned} \tag{53}$$

$$f_{\lambda,\alpha} = 47.7 \cdot \lambda^{-0.0019} - 46.7 \tag{54}$$
$$f_{AR,\alpha} = 1.294 - 0.590 \cdot AR^{-0.662} \tag{55}$$
$$f_{a,a} = 0.776 + 1.911 \cdot Re^{-06876} \tag{56}$$

where $\ddot{\beta}_z$ denotes the projection of the current wing surface angular acceleration on the $Z_{w,i}$ axis; $\ddot{\beta}_y$ represents the projection of the current wing surface angular acceleration on the $Y_{w,i}$ axis; $LE(r)$ is the Z-axis coordinate of the leading edge at the position $r$; $TE(r)$ is the Z-axis coordinate of the trailing edge at the position $r$; $rot(r)$ specifies the Z-axis coordinate of the rotation axis at the position $r$.

Consequently, the aerodynamic force and moment generated due to the added mass effect are:

$$F_{x,add,w,i} = \int_{\Delta R}^{R} \Delta F_{x,add,w,i} \tag{57}$$

$$\begin{bmatrix}T_{x,add,i}\\T_{y,add,i}\\T_{z,add,i}\end{bmatrix} = \begin{bmatrix}0.0\\r_{y,add,w,i}\\r_{z,add,w,i}\end{bmatrix} \times \begin{bmatrix}F_{x,add,w,i}\\F_{y,add,w,i}\\F_{z,add,w,i}\end{bmatrix} = \begin{bmatrix}0.0\\-\Delta F_{x,add,w,i} \cdot L_{C,add,cop}\\-\Delta F_{x,add,w,i} \cdot L_{S,add,cop}\end{bmatrix} \tag{58}$$

$$L_{C,add,cop} = 1.078 \cdot R_2 \tag{59}$$
$$L_{S,add,cop} = 0.500 \cdot \bar{c} \tag{60}$$
$$F_{x,add,w,i} = F_{x,add,w,i} \tag{61}$$
$$F_{y,add,w,i} = 0.0 \tag{62}$$
$$F_{z,add,w,i} = 0.0 \tag{63}$$
$$T_{x,add,i} = 0.0 \tag{64}$$
$$T_{y,add,i} = -\Delta F_{x,add,w,i} \cdot L_{C,add,cop} \tag{65}$$
$$T_{z,add,i} = -\Delta F_{x,add,w,i} \cdot L_{S,add,cop} \tag{66}$$

Within the wing surface coordinate system, the wing experiences tri-axial forces and corresponding moments due to aerodynamic phenomena as follows:

$$F_{x,tol,w,i} = F_{x,tr,w,i} + F_{x,rot,w,i} + F_{x,add,w,i} \tag{67}$$
$$F_{y,tol,w,i} = 0 \tag{68}$$
$$F_{z,tol,w,i} = F_{z,tr,w,i} + F_{z,rot,w,i} + F_{z,add,w,i} \tag{69}$$
$$T_{y,wing,w,i} = T_{y,tr,i} + T_{y,rot,i} + T_{y,add,i} \tag{70}$$

$$T_{z,\text{wing},w,i} = T_{z,tr,i} + T_{z,rot,i} + T_{z,add,i} \tag{71}$$

## 2.4 Tensile modeling of single wings

This section determines the parameters $C_{wing,i}$ and $K_{wing,i}$ within the model mentioned above. As discussed previously, the flexible membrane part of the wing undergoes reciprocating tension at a relatively high frequency during the flapping process. Consequently, the viscoelastic phenomena[32] that arise must be addressed, necessitating consideration of the loading rate's impact on the tensioning process.

In terms of tension condition assessment, this study triggers tension conditions where the absolute values of the torsional angular velocity and the flapping angular velocity are considered, primarily when their motion patterns are inversely related.

$$C_{\text{tens}} = [C_{scale1} \cdot (\dot{\theta} \cdot \dot{\phi}) - C_{move}] \tag{72}$$

$$C_{Tension,\ i} = \frac{C_{c1}}{1 + e^{-C_{\text{tens}}^2}} \tag{73}$$

Here, $\dot{\theta}$ represents the rotation angular velocity, and $\dot{\phi}$ denotes the flapping angular velocity, with $C_{c1}$, $C_{scale1}$ and $C_{move}$ being constant parameters. To determine the stiffness and damping resulting from the viscoelastic phenomena, presented as follows:

$$K_{wing,i} = C_{Tension,\ i} \cdot K_{w-s,i} \tag{74}$$

$$K_{w-s,i} = \frac{k_{w-s,i}}{k_{w-s-ref,i}} \tag{75}$$

$$k_{w-s,i} = ((k_1 - \frac{1}{\sigma\sqrt{2\pi}} e^{-0.5(\frac{\theta_n-\mu}{\sigma})^8})/k_1)^2 \\ - ((k_1 - \frac{1}{\sigma\sqrt{2\pi}} e^{-0.5(\frac{-\mu}{\sigma})^8})/k_1)^2 \tag{76}$$

$$k_{w-s-ref,i} = ((k_1 - \frac{1}{\sigma\sqrt{2\pi}} e^{-0.5(\frac{\theta_{tat}-\mu}{\sigma})^8})/k_1)^2 \\ - ((k_1 - \frac{1}{\sigma\sqrt{2\pi}} e^{-0.5(\frac{-\mu}{\sigma})^8})/k_1)^2 \tag{77}$$

$$\theta_n = k_1 \cdot \theta \tag{78}$$

$$C_{wing,i} = C_{Tension,\ i} \cdot C_{wing-ref,i} \tag{79}$$

With $\theta_{tat}$, $k_1$, $\sigma$, $\mu$, and $C_{wing-ref,i}$ as constant parameters, the model's parameters are selected from a feasible solution set referencing the magnitude of parameters from existing research.

## 2.5 Influence coefficients in tandem wing configurations

This section is dedicated to determining the $C_{tandem,i}$ within the discussed model, which primarily focuses on the nonlinear characteristics of interest in this research. Refer to Appendix C for the detailed process of determining the tandem wing influence coefficients. The final form of the tandem wing influence coefficients is presented as follows:

$$X_0 = \dot{\phi}_f/(w_{maxf} \cdot C_{am}) \tag{80}$$

$$X_1 = \dot{\phi}_h/(w_{maxh} \cdot C_{am}) \tag{81}$$

$$X_2 = (\dot{\phi}_f - \dot{\phi}_h)/(w_{maxd} \cdot C_{am}) \tag{82}$$

$$X_3 = \phi_f/C_{am} \tag{83}$$

$$X_4 = \phi_h/C_{am} \tag{84}$$

$$X_5 = (\phi_f - \phi_h)/C_{am} \tag{85}$$

$$X_6 = \sin(2 \cdot \phi_f \cdot (\pi/C_{am})) \tag{86}$$

$$X_7 = \sin(2 \cdot \phi_h \cdot (\pi/C_{am})) \tag{87}$$

$$X_8 = \sin(4 \cdot \phi_f \cdot (\pi/C_{am})) \tag{88}$$

$$X_9 = \sin(4 \cdot \phi_h \cdot (\pi/C_{am})) \tag{89}$$

$$X_{10} = \sin(8 \cdot \phi_f \cdot (\pi/C_{am})) \tag{90}$$

$$X_{11} = \sin(8 \cdot \phi_h \cdot (\pi/C_{am})) \tag{91}$$

$$\begin{aligned}
C_{i-TF} = &\, X_0 \cdot X_8 \cdot (-11.453 \cdot X_0 \cdot (-5 \cdot X_6 + X_7) - 22.906 \cdot X_1 \\
&+ 11.453 \cdot X_2 - 11.453 \cdot X_4 + 11.453 \cdot X_5 - 11.453 \\
&\cdot X_7 - 11.453 \cdot \sin(\sin(X_8))) + X_0 \cdot X_9 \cdot (11.453 \\
&\cdot X_{10} + 11.453 \cdot X_{11} + 11.453 \cdot X_2 - 22.906 \cdot X_4 \\
&- 11.453 \cdot X_5 \cdot (X_1 - X_{11} - X_5) - 11.453 \cdot X_7) + 9 \\
&\cdot X_1 - 7.892 \cdot X_2 + 7.892 \cdot X_4 + X_8 - 6.166
\end{aligned} \tag{92}$$

$$\begin{aligned}
C_{i-TH} = &\, 49.314 \cdot X_1 - (X_2 - X_5 - \sin(X_1 - 124.935)) \cdot (X_2 + X_4 \\
&- X_8 - 0.994) \cdot (X_3 + X_4 + X_6 + X_7) \cdot (-2X_1 - X_{10} \\
&- X_{11} + X_2 - X_7 \cdot (X_1 + X_7 - \sin(X_{10} - X_5)) \cdot (X_1 \\
&+ X_{11} - X_9 - 45.822) - X_8 - (X_2 + X_4 - \sin(X_1 \\
&- 124.935))(2 \cdot X_0 - 3 \cdot X_{10} - X_9 - 34.855)(X_3 + X_4 \\
&+ X_6 + X_7) - 49.314)
\end{aligned} \tag{93}$$

where $w_{maxf}$ denotes the maximum flapping angular velocity of the Forewing, $w_{maxh}$ represents the maximum flapping angular velocity of the hindwing, and $C_{am}$ is the amplitude of flapping.

## 2.6 Model of motor stiffness

This section is focused on determining the $K_{motor,i}$ within the mentioned model. The objective is to leverage the resonance characteristics[33] to minimize the driving torque of the servo motors. In each experimental state, torsion springs with the following stiffness form are installed to achieve this goal:

$$K_{motor,i} = K_{ref} \cdot f^2 \tag{94}$$

where $f$ denotes the flapping frequency, $K_{ref}$ is a constant.

# 3 Methodology

## 3.1 Architecture of the ConcertoRL algorithm

The architecture of the ConcertoRL algorithm, as illustrated in the accompanying Fig. 6, integrates a time-interleaved mechanism to refine control precision with a rule-based policy composer designed to enhance the stability of the online training.

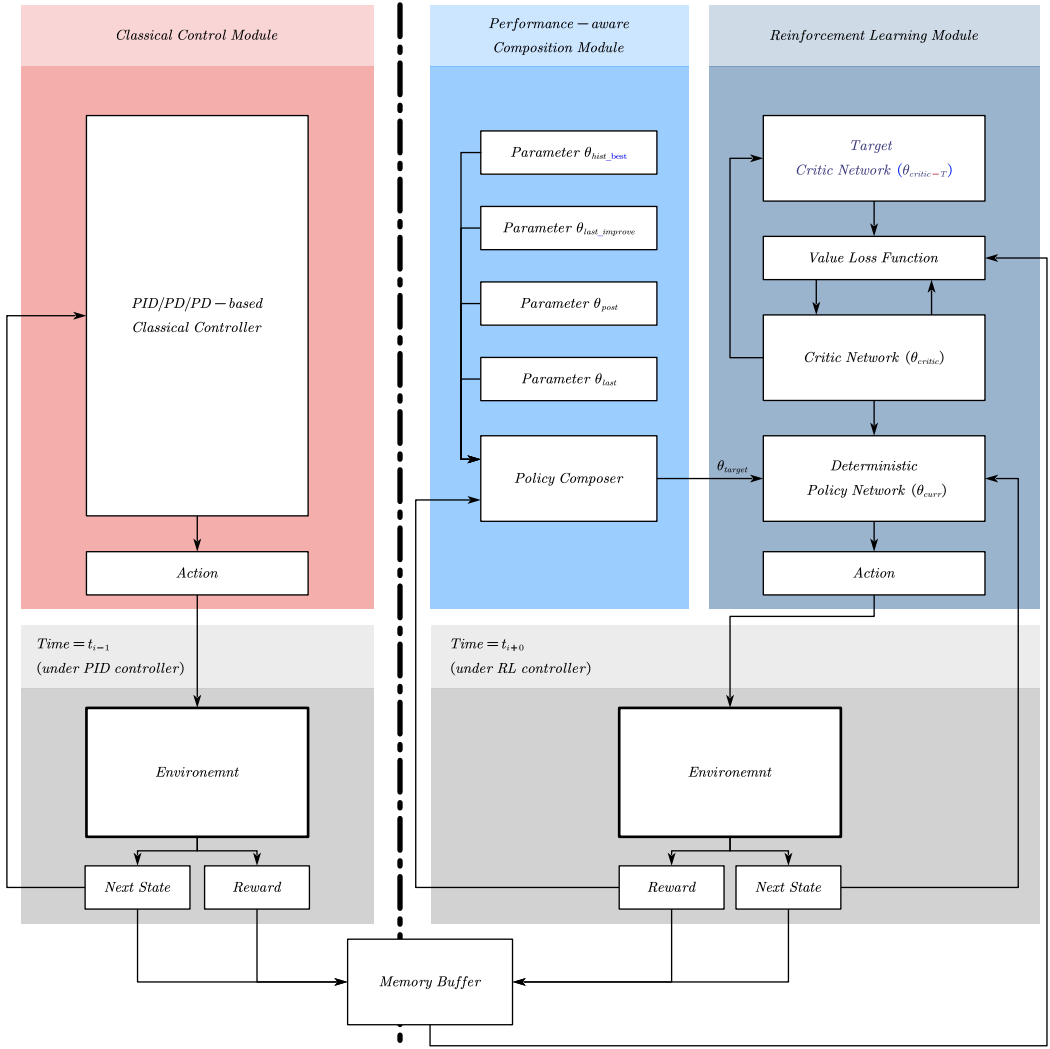

**Fig. 6.** ConcertoRL Algorithm's Architecture

## 3.2 Integration of time-interleaved mechanisms with classical control algorithms

The approach addresses the challenges associated with combining traditional controllers with RL controllers, which have primarily been categorized into several paradigms as follows:

1. **RL for Tuning Classical Controller Parameters[34]:** This approach does not guarantee the stability of the RL exploration process in its initial stages.
2. **Classical Controllers for Guided Policy Search[35]:** The performance ceiling is limited by the capabilities of the classical controller.
3. **Weighted Combination of Classical Controllers and the RL[36]:** This fails to ensure the stability of the RL exploration process.

To address the challenges above, particularly the computational intensity associated with high-frequency RL control in real-time operations and the need for integration with existing control systems, this study introduces Time-Interleaved Mechanisms. The core concept of Time-Interleaved Mechanisms lies in mitigating the adverse effects of suboptimal actions during RL's unfinished training and exploration phases, which is achieved by segmenting the timeline and incorporating proven classical controllers to suppress the negative impacts of RL-induced suboptimal actions. Furthermore, this approach indirectly relocates the exploration process to the realm of traditional control, thereby avoiding the introduction of random noise into the online, continuous learning of nonlinear systems. Such a mechanism facilitates a decoupling and balance between exploration and exploitation in reinforcement learning. The pseudocode for this mechanism is provided below.

```
Algorithm 1: Time-interleaved Mechanism
1  for i_step ← 0 to N do
2      if i_step mod 2 = 0 then
3          action ← agent.select_action(s_t)
4      else
5          A_1 ← pid_controller_1.update_ACT(φ_{EXP-n-1} − φ_{SIM-m-1})
6          A_2 ← pid_controller_2.update_ACT(φ_{EXP-n-2} − φ_{SIM-m-2})
7          A_3 ← pid_controller_3.update_ACT(φ_{EXP-n-3} − φ_{SIM-m-3})
8          A_4 ← pid_controller_4.update_ACT(φ_{EXP-n-4} − φ_{SIM-m-4})
9          action ← ([A_1, A_2, A_3, A_4])
10     end
11 end
```

In selecting classical controllers, as elucidated by the pseudocode, the Time-Interleaved Mechanisms impose no restrictions on the type of classical controllers employed. This paper adopts the widely utilized PID-type controllers[37] to achieve a data-driven approach and facilitate plug-and-play functionality. We have conducted tests on controllers with varying characteristics, including PID, PD, and PI controllers[38, 39], and PID controllers with different parameter settings to evaluate their integration effectiveness within the Time-Interleaved Mechanisms.

## 3.3 Development of a rule-based policy composer

In the context of ongoing online learning processes, it is crucial to circumvent policy adjustments that could lead to deteriorated optimization directions. Common strategies to address this challenge are categorized as follows, with their applicability to the current issue assessed:

---

**Algorithm 2:** Rule-Based Policy Composer

**Input** : $R_{\text{deque}}, V_{\text{hist}}, V_{\text{last}}, C_{\text{unb}}, C_{\text{ref}}, \pi, \pi_{\text{targ}}, \pi_{\text{hist}}, \pi_{\text{lastup}}, \pi_{\text{last}},$
$VS_{\text{last}}, VI_{\text{last}},$

1  $V_{\text{curr}} \leftarrow \text{sum}(R_{\text{deque}})/\text{len}(R_{\text{deque}})$
2  $VS_{\text{curr}}, VI_{\text{curr}} \leftarrow \text{polyfit}(R_{\text{deque}})$
3  **if** $V_{\text{curr}} < V_{\text{hist}}$ **then**
4  $\quad \pi_{\text{targ}} \leftarrow \pi$
5  $\quad \pi_{\text{hist}} \leftarrow \pi$
6  $\quad \pi_{\text{lastup}} \leftarrow \pi$
7  $\quad V_{\text{hist}} \leftarrow V_{\text{curr}}$
8  $\quad V_{\text{last}} \leftarrow V_{\text{curr}}$
9  **else**
10  $\quad C_{\text{sca}} \leftarrow C_{\text{ref}} \times (V_{\text{curr}}/V_{\text{last}})$
11  $\quad$ **if** $V_{\text{curr}} < V_{\text{last}}$ **then**
12  $\quad\quad \pi_{\text{targ}} \leftarrow \pi$
13  $\quad\quad$ **if** $V_{\text{curr}} < V_{\text{hist}} \times 1.05$ **then**
14  $\quad\quad\quad \pi_{\text{lastup}} \leftarrow \pi$
15  $\quad\quad\quad V_{\text{last}} \leftarrow V_{\text{curr}}$
16  $\quad\quad$ **else**
17  $\quad\quad\quad$ **if** $VS_{\text{curr}} < VS_{\text{last}}$ **then**
18  $\quad\quad\quad\quad$ **if** $VI_{\text{curr}} < VI_{\text{last}}$ **then**
19  $\quad\quad\quad\quad\quad \pi_{\text{targ}} \leftarrow \pi$
20  $\quad\quad\quad\quad$ **else**
21  $\quad\quad\quad\quad\quad \pi_{\text{targ}} \leftarrow \pi_{\text{lastup}}$
22  $\quad\quad\quad\quad$ **end**
23  $\quad\quad\quad$ **else**
24  $\quad\quad\quad\quad$ **if** $VI_{\text{curr}} < VI_{\text{last}}$ **then**
25  $\quad\quad\quad\quad\quad \pi_{\text{targ}} \leftarrow \pi * 1 + \pi_{\text{last}} \times (C_{\text{sca}} - 1) + \pi_{\text{lastup}} \times (1 - C_{\text{sca}})$
26  $\quad\quad\quad\quad$ **else**
27  $\quad\quad\quad\quad\quad \pi_{\text{targ}} \leftarrow \pi_{\text{lastup}}$
28  $\quad\quad\quad\quad$ **end**
29  $\quad\quad\quad$ **end**
30  $\quad\quad$ **end**
31  $\quad \pi_{\text{last}} \leftarrow \pi$
32  $\quad \pi \leftarrow \pi_{\text{targ}}$
33  $\quad VI_{\text{last}} \leftarrow VI_{\text{curr}}$
34  $\quad VS_{\text{last}} \leftarrow VS_{\text{curr}}$
35  $\quad V_{\text{last}} \leftarrow V_{\text{curr}}$
36  **end**
37  **return** $\pi, \pi_{\text{hist}}, \pi_{\text{last}}, V_{\text{last}}, V_{\text{hist}}, VI_{\text{last}}, VS_{\text{last}}$

1. **Meta-controllers[40]:** While meta-controllers offer a potential solution, they require training, which may not guarantee effective early performance. Pre-trained models can mitigate this but at the cost of additional computational resources.
2. **Halting experimental apparatus upon bug detection[41]:** This approach precludes the possibility of continuous online experimentation.

In response to these challenges and to assign posterior weights to states from previous iterations, among other considerations, this study proposes a rule-based policy composer. The underlying principle involves periodic linear fitting of the reward curve. Based on the slope and intercept derived from the linear fit, in conjunction with the slope and intercept from the previous fit, average rewards from the previous and current phases, one of six strategies is selected to adjust the network weights of the policy:

## 3.4 Defining the state and action spaces

Building on the analysis presented, the core challenge of the current problem lies in its inherently nonlinear and non-stationary nature, coupled with partial observability issues. Our approach leverages historical data to extract pertinent information and ascertain the characteristics of the current state. Following the existing research methodology[42], our study incorporates multiple past event steps and future commands to encapsulate information about the system's state.

$$s_t = [E_{0+n}, \cdots, E_0, (S_{or-t-1}, A_{t-1}), \cdots, (s_{or-t-m}, a_{t-m})] \tag{95}$$

$$E_{0+n} = [\phi_{EXP-n-1}, \phi_{EXP-n-2}, \phi_{EXP-n-3}, \phi_{EXP-n-4}] \tag{96}$$

$$S_{or-t-m} = [\phi_{SIM-m-1}, \phi_{SIM-m-2}, \phi_{SIM-m-3}, \phi_{SIM-m-4}] \tag{97}$$

where $\phi_{EXP-n-i}$ denotes the expected flapping angle position of the i-th wing for the next n steps; $\phi_{SIM-m-i}$ represents the actual flapping angle position of the i-th wing for the past m steps; $a_{t-j}$ indicates the action taken at the j-th step prior.

## 3.5 Designing reward functions and simulation episodes

In this study, the reward function is articulated as a weighted sum of the absolute values of tracking errors for four motors:

$$R_t(s_t, a_t) = \lambda \cdot [|E_1(s_t, a_t)| + |E_2(s_t, a_t)| + |E_3(s_t, a_t)| + |E_4(s_t, a_t)|] \tag{98}$$

where $E_1(s_t, a_t)$, $E_2(s_t, a_t)$, $E_3(s_t, a_t)$, $E_4(s_t, a_t)$ denote the tracking errors of the four motors. $\lambda$ is weighting factors.

To calculate a cumulative reward over discrete phases within the study, the

formulation for the average reward is proposed as follows:

$$V_{t\sim t+m} = \frac{\sum_{i=0}^{m} R_{t+i}}{m} \tag{99}$$

where $m$ is the requisite number of time steps for averaging.

Within the domain of Simulator Episode Design, any deviation exceeding 90 degrees is presumed to inflict damage on the equipment, emphasizing the criticality of precision in actuator control.

## 3.6 Constructing the policy network

The deliberate design of network architecture plays a crucial role in hastening convergence. The prevalent actor network structures are examined below, alongside their limitations:

1. **Standard Multilayer Perceptron (MLP)**: Demonstrates a comparatively weak ability to leverage sequential state information.
2. **Standard Recurrent Neural Network (RNN)**: Offers improved capabilities in processing time-series data; however, its reliance on hidden states makes training difficult and limits its applicability for inference and training on low-power hardware.

In response to these challenges, our research has developed a policy network structure based on tensor subtraction and multiplication, as shown in Fig. 7. This design strategy is aimed at maximizing the network's expressive power using the fewest possible parameters and enhancing the network's utilization of time-series features especially the differentiation between states without the need for RNNs.

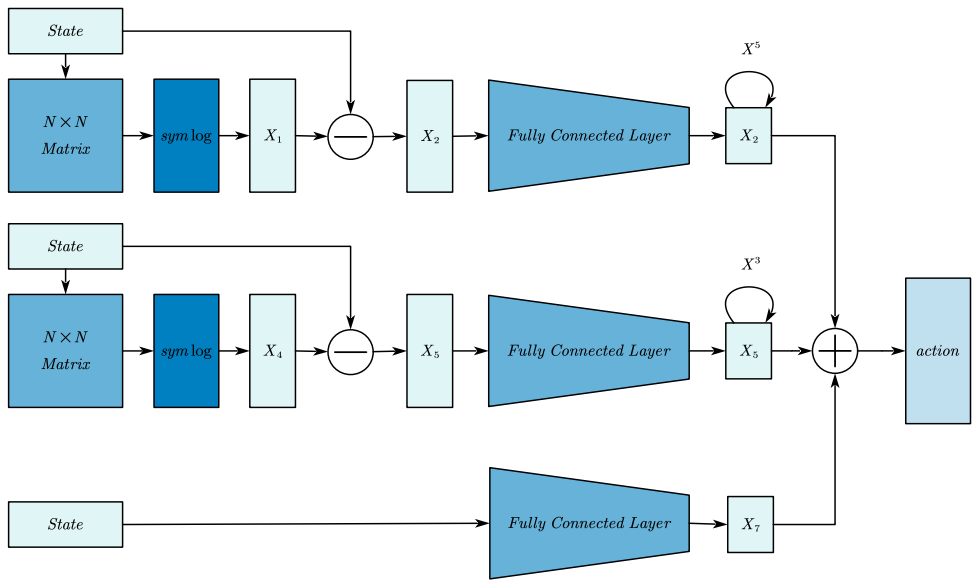

**Fig. 7.** Network structure

# 4 Experiments

## 4.1 Experimental settings

All network parameters were optimized using the Lion optimizer[43], with learning rates set at 0.015 for the actor and 0.0015 for the critic. The batch size was determined to be 128, while the discount factor and the soft update parameter were set at 0.9 and 0.05, respectively. The iteration count was chosen to be 450,000 steps, corresponding to a time step of 0.0005 seconds.

In the following simulation, the following four working conditions are selected as shown in table 1:

**Table 1** Selected operating conditions

| Condition | $f$ | $A_1$ | $A_2$ | $A_3$ | $A_4$ |
|---|---|---|---|---|---|
| 20Hz Standard | 20 | 60° | 60° | 60° | 60° |
| 40Hz Standard | 40 | 60° | 60° | 60° | 60° |
| 40Hz Maneuvering | 40 | 40° | 80° | 40° | 80° |
| 60Hz Standard | 60 | 60° | 60° | 60° | 60° |

## 4.2 Challenges with SAC algorithm

This section explores the performance evaluation of the SAC algorithm across various operational scenarios, concentrating on metrics such as average reward, reward efficiency, and the tracking accuracy of four servo motors. The aim was to unearth and scrutinize the intrinsic challenges encountered by classical reinforcement learning algorithms like SAC when tackling the specific problem formulated in this investigation.

Under the 20Hz Standard Condition, the SAC algorithm's performance was meticulously evaluated using criteria including average reward, total reward, and the tracking precision of four servo motors. The outcomes, systematically depicted in Fig. 8, Fig. 9, and Fig. 10, reveal that due to SAC's exploration strategy and its incomplete training phase, there was a general trend of significant tracking errors early in the training, often reaching a 90-degree error threshold, which consequently met the episode termination condition due to potential equipment damage.

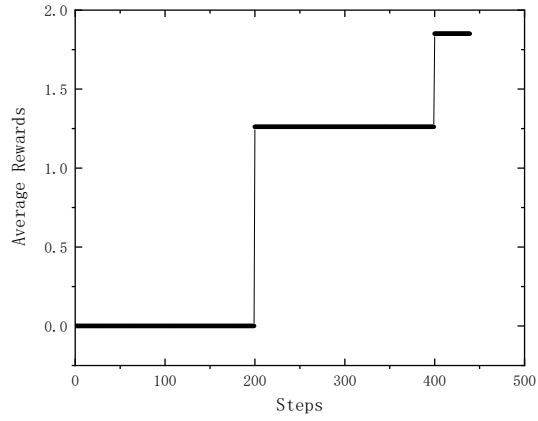

**Fig. 8.** Average reward under 20Hz standard condition

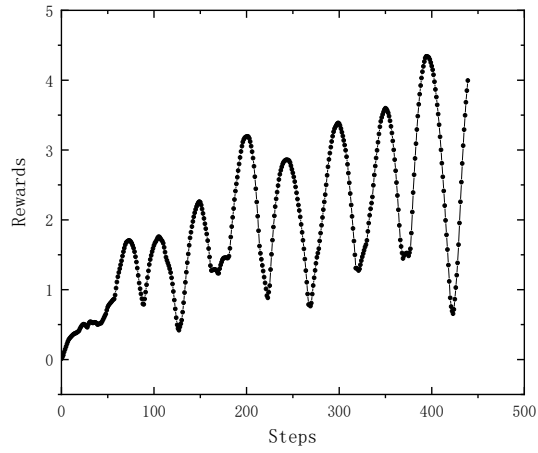

**Fig. 9.** Reward under 20Hz standard condition

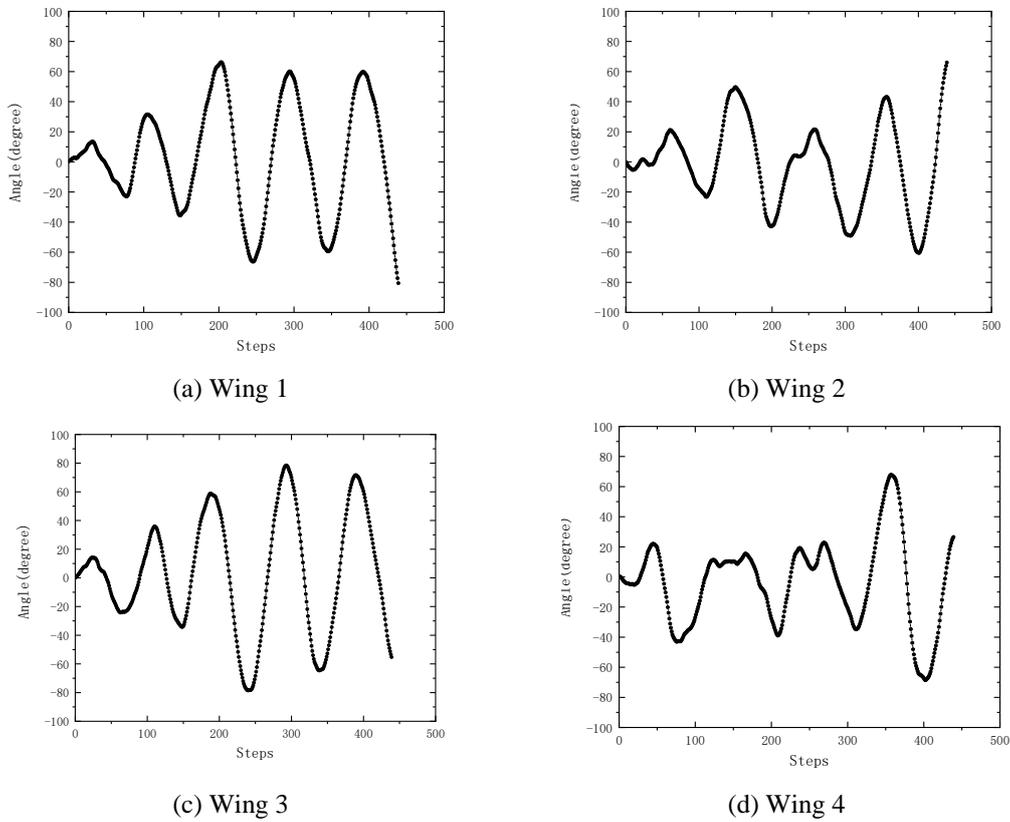

(a) Wing 1      (b) Wing 2

(c) Wing 3      (d) Wing 4

**Fig. 10.** Tracking error under 20Hz standard condition

In the 40Hz Standard Condition, with increased unsteadiness and nonlinearity of the problem, a similar set of metrics was employed to assess the SAC algorithm, and the results are shown in Fig. 11, Fig. 12, and Fig. 13. Comparable to the 20Hz Standard Condition, the algorithm rapidly reached the episode termination condition within a relatively short period.

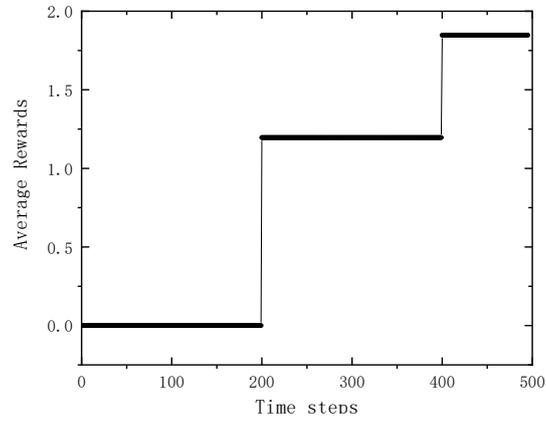

**Fig. 11.**  Average reward under40Hz standard condition

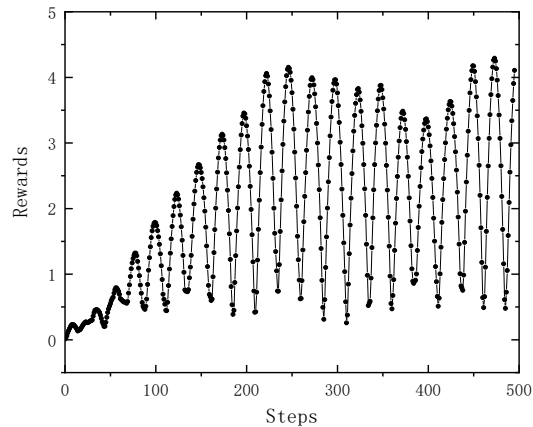

**Fig. 12.**  Reward under 40Hz standard condition

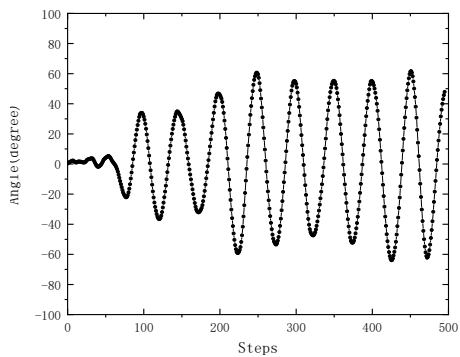 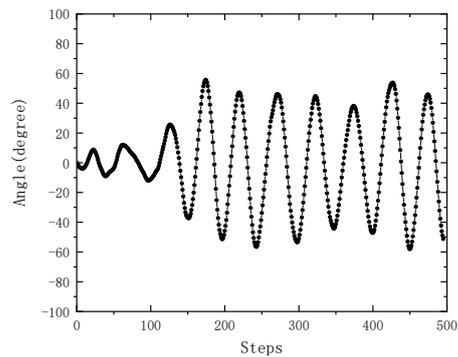

(a) Wing 1　　　　　　　　　　　　(b) Wing 2

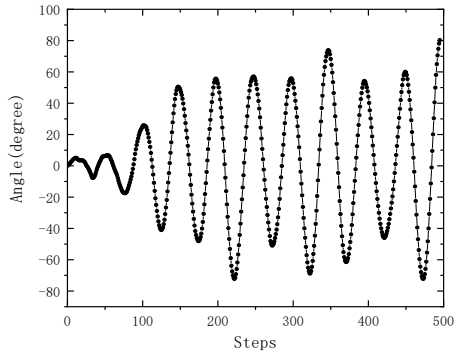
(c) Wing 3

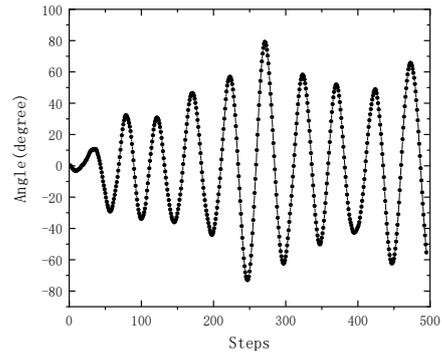
(d) Wing 4

**Fig. 13.** Tracking error under 40Hz standard condition

The 40Hz Maneuvering Condition posed a more challenging scenario for the SAC algorithm, testing its adaptability and control under dynamic conditions. Fig. 14, Fig. 15, and Fig. 16 showcase the average reward, total reward, and servo motor tracking performance in this setting, indicating that under the current parameter configuration, the termination condition was met in less than 250 steps due to the same issue of equipment damage.

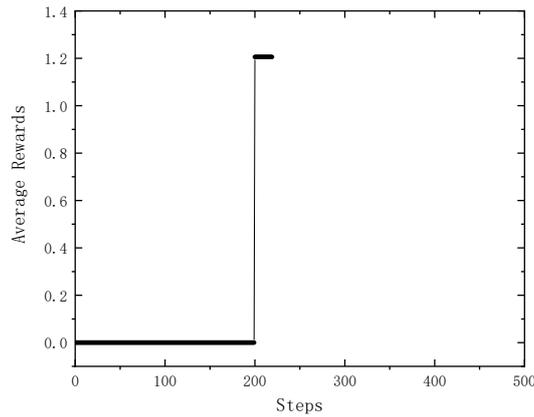

**Fig. 14.** Average reward under 40Hz maneuvering condition

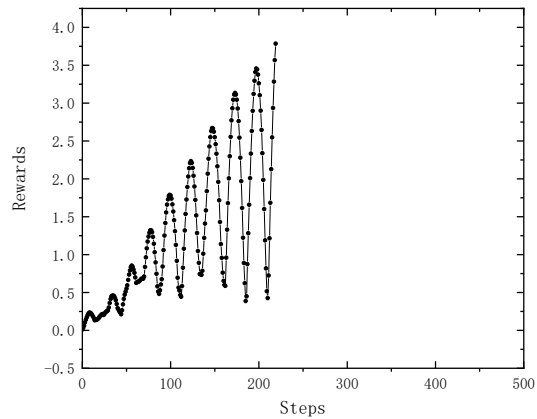

**Fig. 15.** Reward under 40Hz maneuvering condition

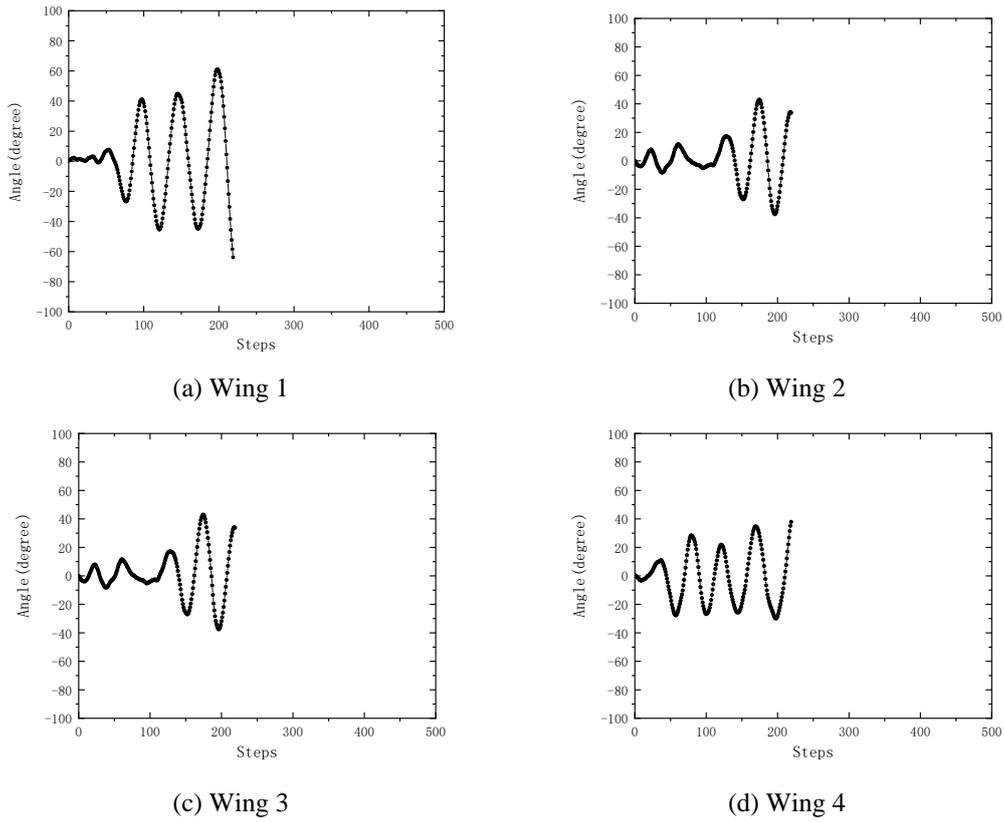

(a) Wing 1  (b) Wing 2
(c) Wing 3  (d) Wing 4

**Fig. 16.** Tracking error under 40Hz maneuvering condition

Lastly, the algorithm's performance was examined under the 60Hz Standard Condition. This scenario pushed the SAC algorithm to its limits regarding frequency response and control precision, reaching the maximum experimental scope required. The results are shown in Fig. 17, Fig. 18, and Fig. 19, and similarly, the episode termination condition was encountered due to equipment damage within approximately the first 500 steps, akin to the phenomena observed in previous experiments.

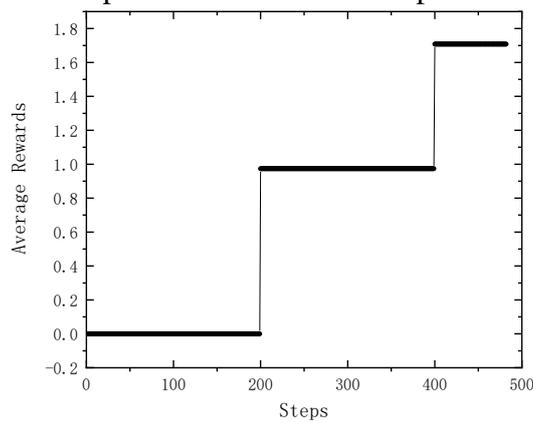

**Fig. 17.** Average reward under 60Hz standard condition

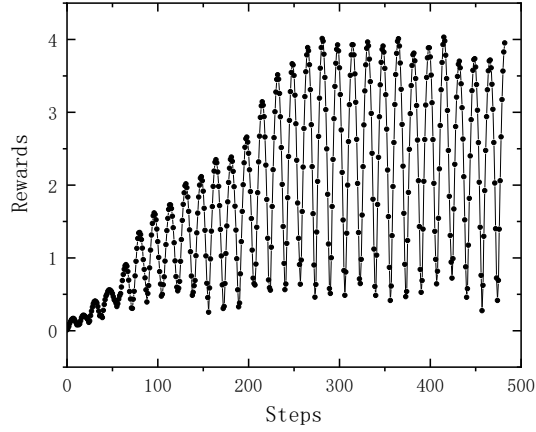

**Fig. 18.** Reward under 60Hz standard condition

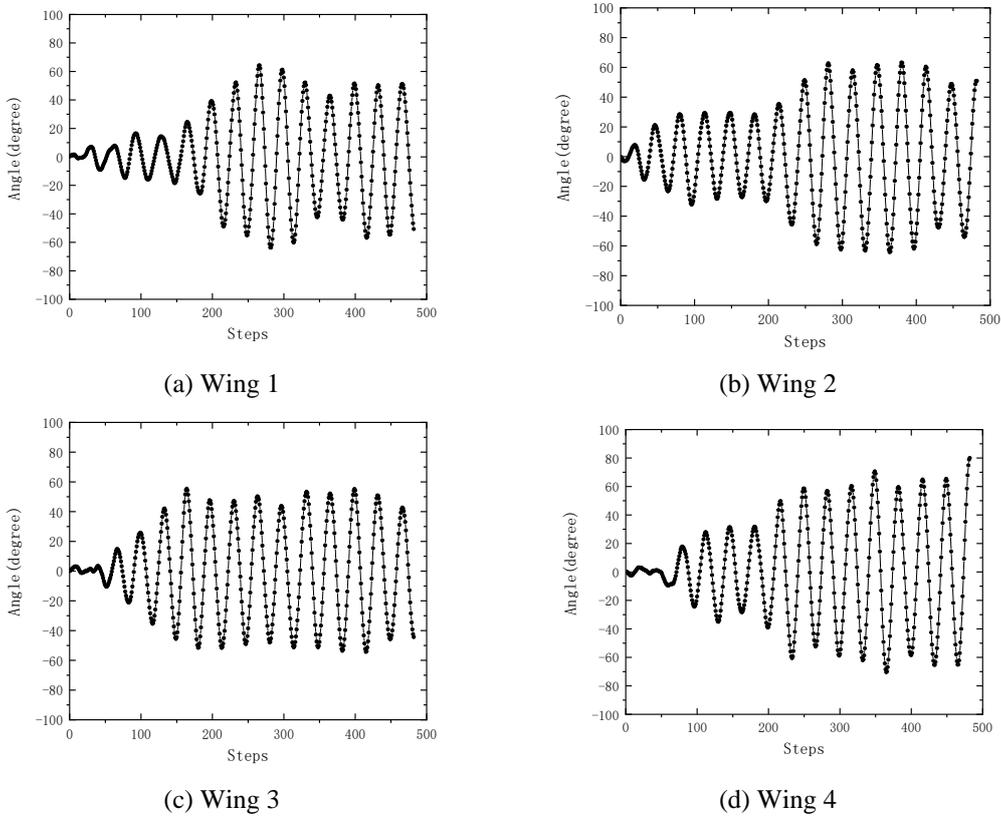

(a) Wing 1     (b) Wing 2

(c) Wing 3     (d) Wing 4

**Fig. 19.** Tracking error under 60Hz standard condition

The observations from these four sets of experiments highlight that while the SAC algorithm has achieved commendable results in numerous control problems, its application in the non-stationary and highly nonlinear control issues presented in this study, coupled with an inadequately trained policy and entropy-based exploration process, leads to rapid termination of episodes due to equipment damage within the initial steps, without showing signs of convergence , which underlines the critical need for developing a high-precision and stable training methodology for reinforcement learning algorithms, as addressed in this paper.

## 4.3 Ablation study on time-interleaved mechanisms

To investigate the characteristics of the time-interleaved mechanisms, the following three control experiments are designed:

1. **1000Hz PID**: This setting does not include Time-Interleaved Mechanisms, serving as the baseline for comparison.
2. **2000Hz PID**: Incorporates Time-Interleaved Mechanisms with all controllers being PID controllers, aimed at analyzing the impact of solely increasing control frequency and as a means to assess if the current algorithm surpasses the performance of traditional controllers.
3. **1000Hz PID+RL**: Features the complete implementation of Time-Interleaved Mechanisms, representing the primary experimental focus.

This section introduces control groups to validate the algorithm's effectiveness under four experimental conditions: 20Hz Standard Condition, 40Hz Standard Condition, 40Hz Maneuvering Condition, and 60Hz Standard Condition. The experiments primarily aim to validate the effectiveness of the reinforcement learning algorithm. In all experiments, the parameters of the PID controllers are kept constant, as shown in Table 2, and the initial weights of the neural networks are identical and aligned with those used in the SAC algorithm. Given the non-learning characteristics of the PID controllers, the performance results at 50,000 iterations serve as the reference values.

**Table 2** PID parameter description

| Parameter | Value |
| --- | --- |
| P | $4.80 \times 10^{-1}$ |
| I | $2.00 \times 10^{-5}$ |
| D | $7.00 \times 10^{-4}$ |

**Table 3** Summary of experimental results

| Condition | Experiments | Best average reward | Performance improvement compared to 1000Hz PID controller | Performance improvement compared to 2000Hz PID controller |
| --- | --- | --- | --- | --- |
| 20Hz Standard | 1000Hz PID | 0.0728 | / | / |
| 20Hz Standard | 2000Hz PID | 0.0389 | / | / |
| 20Hz Standard | 1000Hz PID+RL | 0.0265 | +63.60% | +31.88% |
| 40Hz Standard | 1000Hz PID | 0.217 | / | / |
| 40Hz Standard | 2000Hz PID | 0.136 | / | / |
| 40Hz Standard | 1000Hz PID+RL | 0.0600 | +72.35% | +55.88% |
| 40Hz Maneuvering | 1000Hz PID | 0.233 | / | / |
| 40Hz Maneuvering | 2000Hz PID | 0.146 | / | / |
| 40Hz Maneuvering | 1000Hz PID+RL | 0.0563 | +75.84% | +61.44% |
| 60Hz Standard | 1000Hz PID | 0.343 | / | / |
| 60Hz Standard | 2000Hz PID | 0.246 | / | / |
| 60Hz Standard | 1000Hz PID+RL | 0.0883 | +74.26% | +64.11% |

**Fig. 20.** Average reward under 20Hz standard condition

**Fig. 21.** Reward under 20Hz standard condition

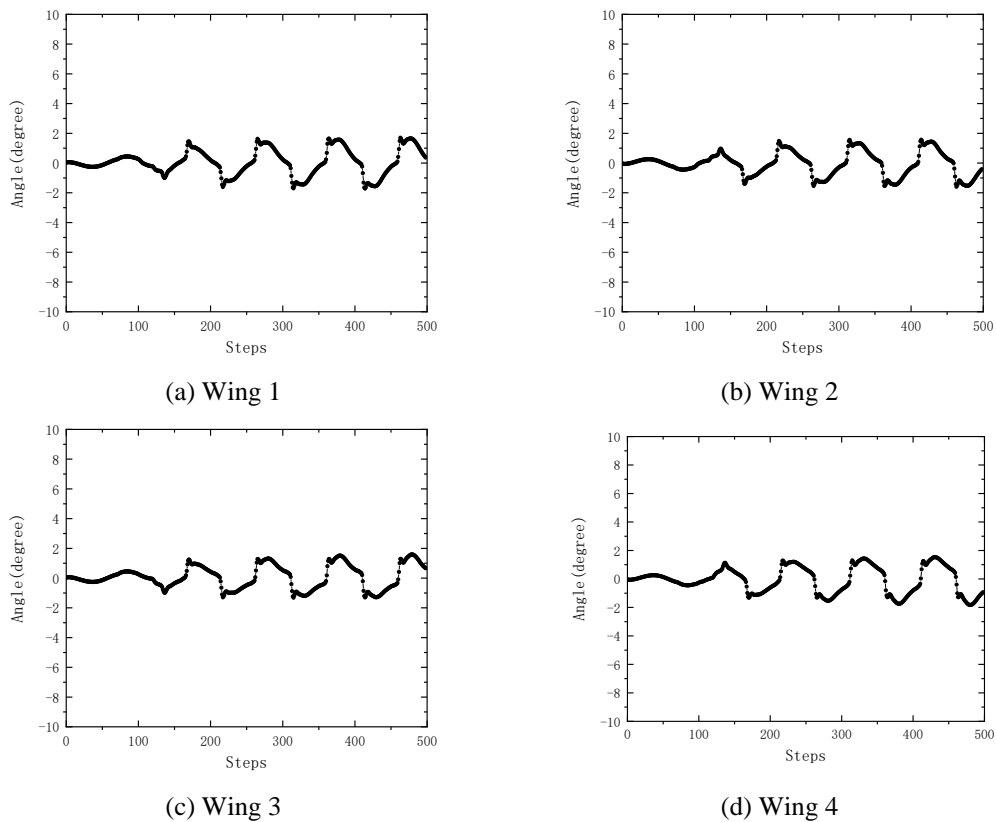

(a) Wing 1   (b) Wing 2

(c) Wing 3   (d) Wing 4

**Fig. 22.** Tracking error of the first 500 steps under 20Hz standard condition

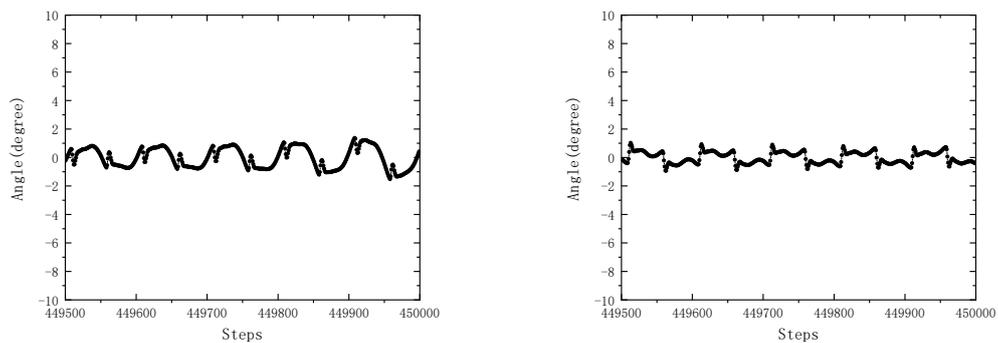

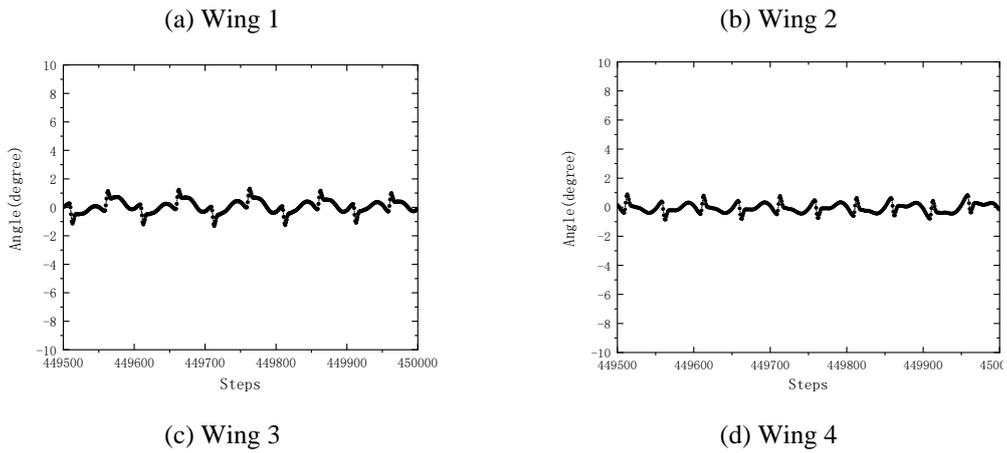

(a) Wing 1　　　　　　　　　(b) Wing 2

(c) Wing 3　　　　　　　　　(d) Wing 4

**Fig. 23.**　Tracking error of the last 500 steps under 20Hz standard condition

**Fig. 24.**　Average reward under 40Hz standard condition

**Fig. 25.**　Reward under 40Hz standard condition

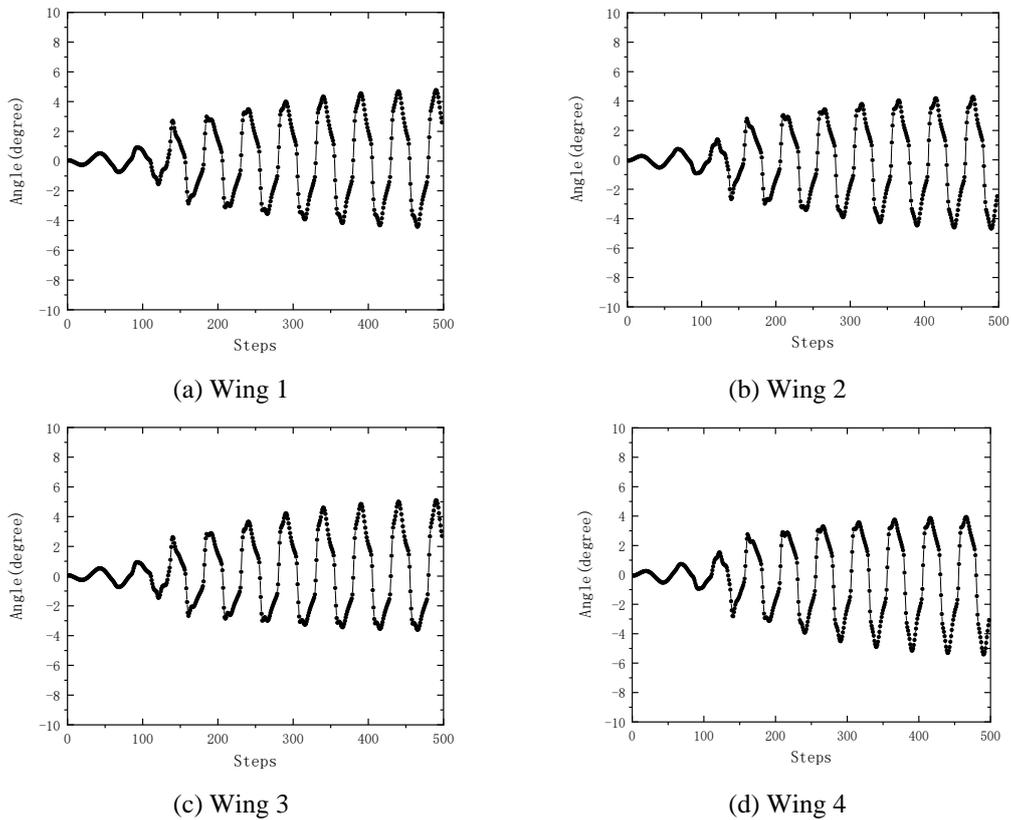

(a) Wing 1　　　　　　　　　(b) Wing 2

(c) Wing 3　　　　　　　　　(d) Wing 4

**Fig. 26.**　Tracking error of the first 500 steps under 40Hz standard condition

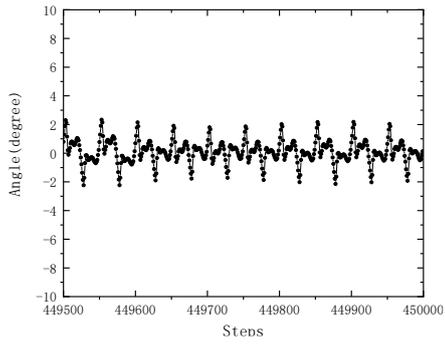

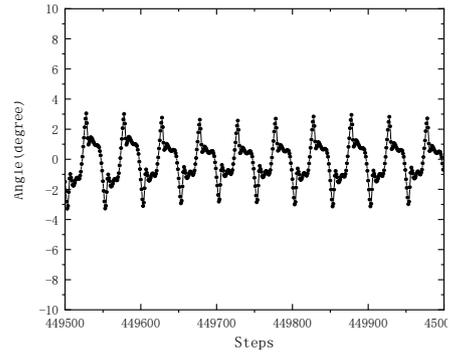

(a) Wing 1

(b) Wing 2

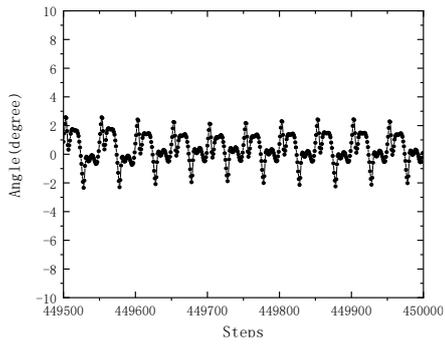

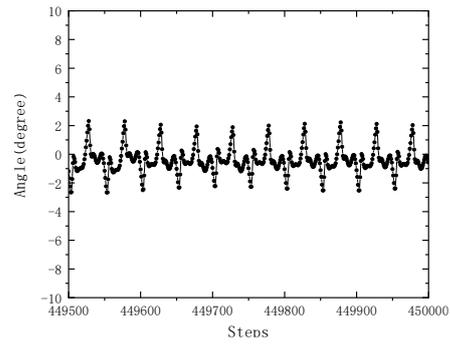

(c) Wing 3

(d) Wing 4

**Fig. 27.** Tracking error of the last 500 steps under 40Hz standard condition

**Fig. 28.** Average reward under 40Hz maneuvering condition

**Fig. 29.** Reward under 40Hz maneuvering condition

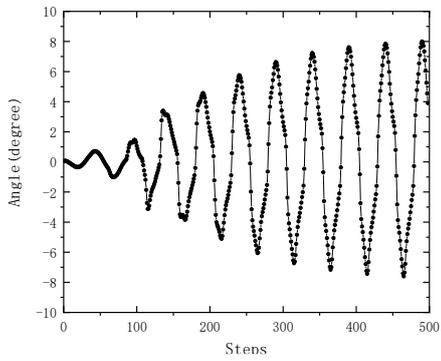

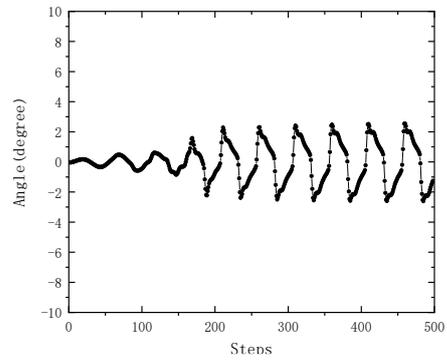

(a) Wing 1

(b) Wing 2

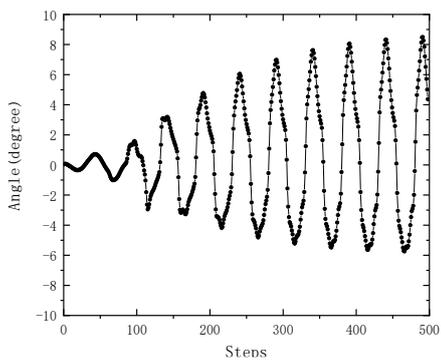

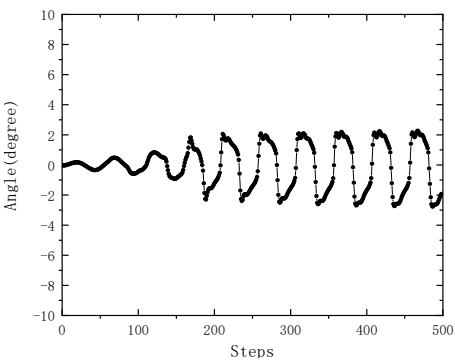

(c) Wing 3           (d) Wing 4

**Fig. 30.** Tracking error of the first 500 steps under 40Hz maneuvering condition

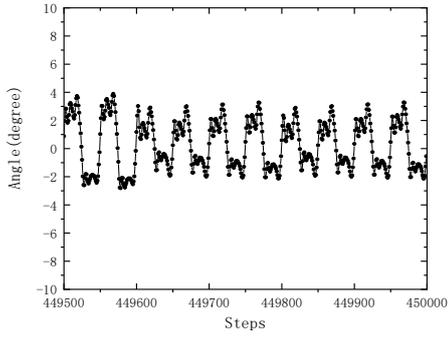
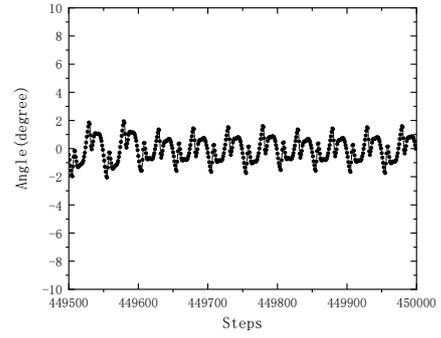

(a) Wing 1           (b) Wing 2

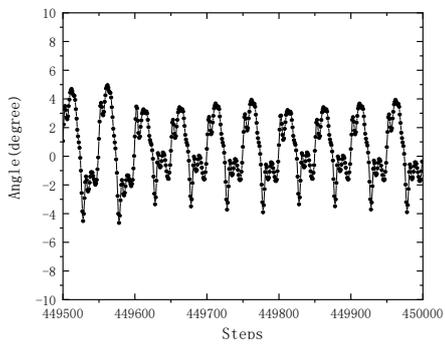
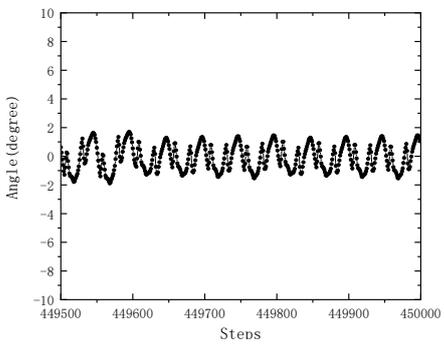

(c) Wing 3           (d) Wing 4

**Fig. 31.** Tracking error of the last 500 steps under 40Hz maneuvering condition

**Fig. 32.** Average reward under 60Hz standard condition

**Fig. 33.** Reward under 60Hz standard condition

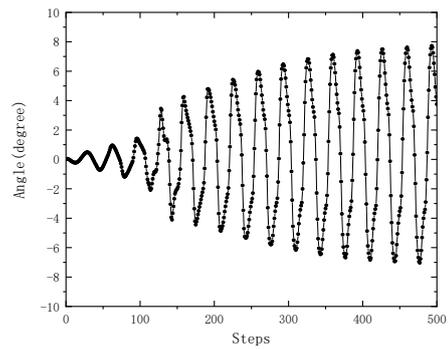
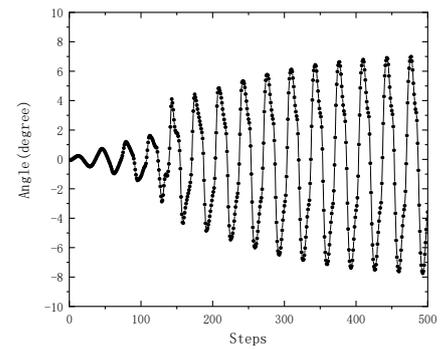

(a) Wing 1           (b) Wing 2

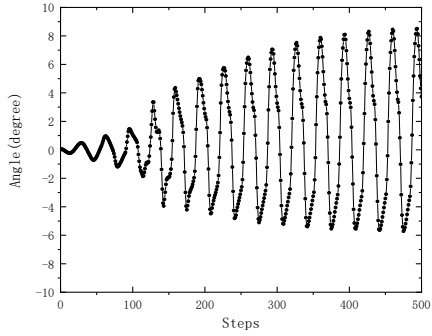
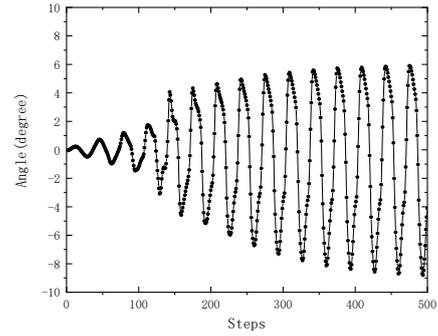

(c) Wing 3  (d) Wing 4

**Fig. 34.** Tracking error of the first 500 steps under 60Hz standard condition

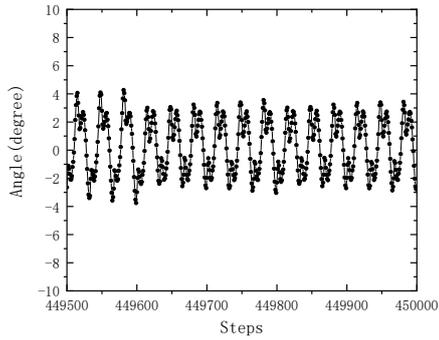
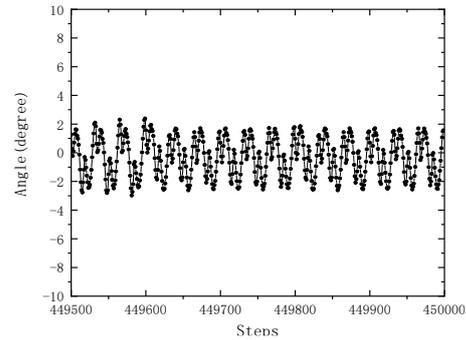

(a) Wing 1  (b) Wing 2

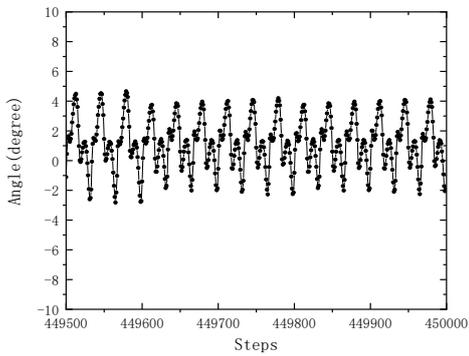
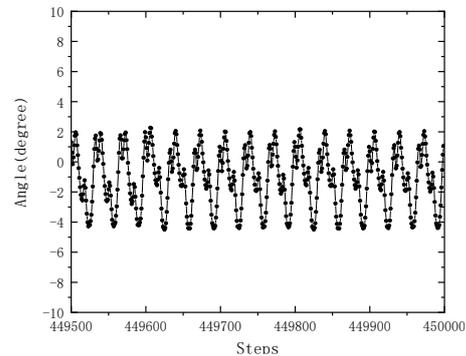

(c) Wing 3  (d) Wing 4

**Fig.35.** Tracking error of the last 500 steps under 60Hz standard condition

According to the above results shown in Table 3 and Figures from Fig.20 to Fig.35, the conclusions drawn from the experiments are as follows:

1. The introduction of Time-Interleaved Mechanisms has significantly enhanced performance. Under the first 500 data points condition, the tracking error of the ConcertoRL algorithm is approximately 10 degrees, representing an order of magnitude improvement over the standard SAC's tracking error of about 90 degrees.

2. The tracking error differences between wing1 and wing14 and wing2 and wing3 indicate that tandem wing interference creates significant operational disparities between the forewing and hindwings. However, ConcertoRL achieves commendable control performance across all four challenging conditions examined in this study, demonstrating that the current ConcertoRL algorithm can address the nonlinear and non-stationary control challenges encountered in the experiment platform.

3. The implementation of Time-Interleaved Mechanisms resulted in over a 60% improvement in tracking error with the ConcertoRL algorithm compared to the 1000Hz PID performance. Furthermore, the tracking error of ConcertoRL also shows more than a 30% improvement over the 2000Hz PID performance. These results indicate that the ConcertoRL algorithm's performance surpasses the upper-performance limits of reference PID controllers, suggesting that ConcertoRL achieves a synergistic effect beyond simple addition. The primary reason for this improvement is its actor-critic architecture-based enhancement of the PID's initial state, thereby improving the overall system performance rather than merely replicating the characteristics of PID controllers as in Guided Policy Search.

## 4.4 Ablation study on rule-based policy composer

This section is dedicated to validating the capability of the proposed rule-based policy composer to prevent the significant degradation of system performance due to the policy weights settling into erroneous local minima during the online continuous training process. An ablation experiment was conducted to assess the effectiveness of this approach.

**Fig. 36.**  Average reward under 20Hz standard condition

**Fig. 36.**  Average reward under 40Hz standard condition

**Fig. 38.**  Average reward under 40Hz maneuvering condition

**Fig. 39.**  Average reward under 60Hz standard condition

According to the above results shown in Table 3 and Figures from Fig.36 to Fig.39, the conclusions drawn from the experiments are as follows:
1. At the onset of online training, the performance of algorithms with and without the rule-based policy composer was similar.
2. Throughout the online training phase, both the algorithm incorporating the rule-based policy composer and the one without it exhibited a general trend of performance improvement. However, the algorithm's performance without the rule-based policy composer significantly deteriorated after a certain period of iterations. This decline is attributed to the emergence of incorrect network weights in the policy, which, given the non-linearity of the problem at hand, led to substantial performance drops due to coupling effects. On the other hand, the algorithm with the rule-based policy composer showed a more stable and consistent improvement trend throughout the process. This stability is due to its mechanism of regressing to previously identified update points with positive trends and continually exploring more effective directions for improvement, thereby achieving sustained performance enhancement.

## 4.5 Universality study under classical controller influence

In the simulation process described in this paper, three primary entities are involved: the control subject, the benchmark control algorithm exemplified by PID controllers, and the reinforcement learning algorithm. To further analyze the performance of the ConcertoRL algorithm and its coupling relationship with classical controllers' performance following the studies above, this section introduces a comparison based on integrating 1000Hz PI controllers, 1000Hz PD controllers, and 1000Hz PID controllers—with parameters randomly adjusted within a range of -10% to +25% by 25% into the ConcertoRL algorithm.

Initially, according to the results shown in Table 4 and Figures from Fig.40 to Fig.43, by combining the PI and PD controllers with the ConcertoRL algorithm, it is observed that the ConcertoRL algorithm still achieves a performance improvement of over 40% across all four experimental conditions, compared to scenarios without the integration of 1000Hz PI or PD controllers, which indicates that the ConcertoRL algorithm maintains algorithmic integrity when combined with controllers exhibiting characteristics different from PID controllers, thereby demonstrating the robust applicability of the ConcertoRL algorithm.

**Table 4** Summary of final average reward

| Condition | classical controller type | final average reward of classical controller | final average reward of RL controller | Performance improvement percentage |
|---|---|---|---|---|
| 20Hz Standard | PI | 0.191 | 0.109 | +42.93% |
| 20Hz Standard | PD | 0.0728 | 0.0372 | +48.90% |
| 40Hz Standard | PI | 0.283 | 0.136 | +51.94% |
| 40Hz Standard | PD | 0.217 | 0.0481 | +77.83% |
| 40Hz Maneuvering | PI | 0.329 | 0.153 | +53.50% |
| 40Hz Maneuvering | PD | 0.233 | 0.0613 | +73.69% |
| 60Hz Standard | PI | 0.586 | 0.161 | +72.53% |
| 60Hz Standard | PD | 0.343 | 0.109 | +68.22% |

(a) PI controller      (b) PD controller

**Fig. 40.** Average reward of experiments under 20Hz standard condition under ConcertoRL algorithm combined with classical controller

(a) PI controller      (b) PD controller

**Fig. 41.** Average reward of experiments under 40Hz standard condition under ConcertoRL algorithm combined with classical controller

(a) PI controller      (b) PD controller

**Fig. 42.** Average reward of experiments under 40Hz maneuvering condition under ConcertoRL algorithm combined with classical controller

(a) PI controller      (b) PD controller

**Fig. 43.** Average reward of experiments under 60Hz standard condition under ConcertoRL algorithm combined with classical controller

Furthermore, according to the results shown in the figures from Fig.44 to Fig.47, when the ConcertoRL algorithm is combined with 1000Hz PID controllers, whose parameters are randomly adjusted within a range of -10% to +25% by 25%, it still facilitates performance improvement under these conditions. This further validates the ConcertoRL algorithm's adaptability and universality.

Notably, in some scenarios, introducing noise can even enhance the performance of ConcertoRL (e.g., under 20Hz standard and 60Hz standard conditions). This improvement could be attributed to the algorithm's consideration of the control subject's nonlinearity and non-stationarity despite the lack of an explicit exploration strategy. The introduction of random noise may enable a degree of direct exploration, which, while potentially leading to performance degradation when combined with nonlinearities (as observed in the 40Hz maneuvering condition), suggests that the decision to utilize randomness for exploration in the face of nonlinear solid challenges like tandem wing interference requires careful consideration.

**Fig. 44.** Average reward under 20Hz standard condition for ConcertoRL algorithm integrating

reference disturbance PID controller and ConcertoRL algorithm integrating constant PID parameters

**Fig. 45.** Average reward under 40Hz standard condition for ConcertoRL algorithm integrating reference disturbance PID controller and ConcertoRL algorithm integrating constant PID parameters

**Fig. 46.** Average reward under 40Hz maneuvering condition for ConcertoRL algorithm integrating reference disturbance PID controller and ConcertoRL algorithm integrating constant PID parameters

**Fig. 47.** Average reward under 60Hz standard condition for ConcertoRL algorithm integrating reference disturbance PID controller and ConcertoRL algorithm integrating constant PID parameters

# 5 Conclusion

This study presented a comprehensive evaluation of the ConcertoRL algorithm, showcasing its effectiveness in addressing nonlinear and non-stationary control challenges through a series of ablation studies and universality assessments under classical controller influence. The findings from our research offer significant insights into the potential of integrating advanced reinforcement learning algorithms with classical control approaches for enhanced system performance.

Implementing Time-Interleaved Mechanisms within the ConcertoRL framework marked a substantial advancement in control accuracy. Specifically, in the early stages of evaluation, the algorithm demonstrated an order of magnitude improvement in tracking error reduction, from approximately 90 degrees in standard SAC implementations to about 10 degrees. This remarkable enhancement underscores the effectiveness of our proposed mechanism in significantly boosting system responsiveness and precision.

Our study further explored the ConcertoRL algorithm's robustness against tandem wing interference, a familiar yet challenging scenario in control systems. The algorithm addressed the significant operational disparities between forewings and hindwings and consistently outperformed traditional PID controllers across all tested conditions. This achievement highlights the ConcertoRL algorithm's capacity to adapt and excel in complex, dynamic environments, validating its applicability and reliability.

Through an ablation study on the rule-based policy composer, we observed a crucial aspect of the ConcertoRL algorithm: its ability to maintain performance improvement trends over extended periods of online training. Unlike approaches without this component, which experienced significant performance degradation due to incorrect policy weight adjustments, the rule-based policy composer ensured a stable and continuous enhancement in system performance. This stability is pivotal for long-term deployment in real-world applications, where system reliability is paramount.

The universality study under classical controller influence revealed that the ConcertoRL algorithm could seamlessly integrate with and improve the performance of traditional PI and PD controllers. Moreover, the algorithm's adaptability was further evidenced by its capacity to benefit from controlled parameter variations and even random noise introductions in specific scenarios. This versatility indicates the algorithm's potential to transcend traditional control paradigms, offering a flexible solution adaptable to various control challenges.

In conclusion, the ConcertoRL algorithm emerges as a highly versatile and effective solution for complex control problems, bridging the gap between reinforcement learning and classical control methodologies. Its demonstrated capability to enhance performance, ensure stability, and maintain adaptability across varied operational conditions positions it as a significant contribution to neural networks and learning systems. Our findings not only validate the efficacy of the ConcertoRL algorithm but also pave the way for future research into hybrid control strategies, promising further advancements in automated and intelligent control systems.

## Acknowledgements

Data will be made available on reasonable request.

## Conflict of interests

# Appendix A  Geometric parameters

The parameters are set as follows.

**Table 5** System parameters

| Parameter | Value and Unit | Value Description |
|---|---|---|
| $D_{X\_B\_FW}$ | + 0.050 m | Select according to the existing experimental bench plan |
| $D_{Y\_B\_FW}$ | + 0.025 m | Select according to the existing experimental bench plan |
| $D_{Z\_B\_FW}$ | + 0.030 m | Select according to the existing experimental bench plan |
| $D_{X\_B\_HW}$ | - 0.050 m | Select according to the existing experimental bench plan |
| $D_{Y\_B\_HW}$ | + 0.025 m | Select according to the existing experimental bench plan |
| $D_{Z\_B\_HW}$ | - 30 / 1000 | Select according to the existing experimental bench plan |
| $J_{ZZ,motor}$ | 0.000000497 kg·m² | Select according to the motor |
| $J_{XX,wing}$ | 0.00000027 kg·m² | Based on Solidworks model selection |
| $J_{YY,wing}$ | 0.00000003 kg·m² | Based on Solidworks model selection |
| $J_{ZZ,wing}$ | 0.00000024 kg·m² | Based on Solidworks model selection |
| $J_{XY,wing}$ | 0.0 kg·m² | Based on Solidworks model selection |
| $J_{XZ,wing}$ | 0.0 kg·m² | Based on Solidworks model selection |
| $J_{YZ,wing}$ | 0.00000006 kg·m² | Based on Solidworks model selection |

# Appendix B  Precision study of the single wing aerodynamic model

This section discusses the experimental model based on the water tunnel experiments conducted by Lua et al. in 2010[44], using the moth wing as a reference. The motion analyzed includes the flapping motion at the wing root and the twisting motion around the leading edge, following a specific flapping pattern as outlined and shown in Fig. 48:

$$\phi_{w,i} = \phi_{am} \cdot cos(2\pi \cdot f \cdot t) \tag{100}$$

$$\theta_{w,i} = \theta_{am} \cdot sin(2\pi \cdot f \cdot t) \tag{101}$$

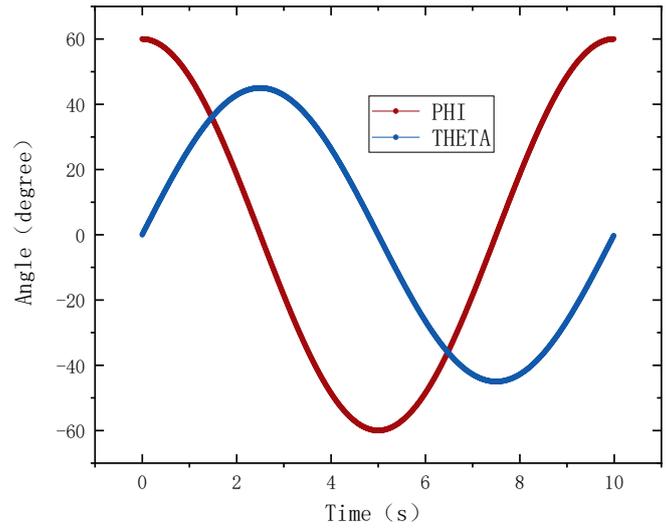

**Fig. 48.** The changing rules of flapping angle and rotation angle selected in the experiment

Given the limitations of the quasi-steady framework, experiments focus on the test rig's validation, selection, and effectiveness of the controller, which revolves around peak accuracy and trend similarity based on the Pearson correlation coefficient. As shown in table 6 and Fig. 49, it is observed that the calculated forces exhibit a trend similar to experimental values, with comparable peak values.

**Table 6** Verification result statistics

| Conditions | Peak load accuracy | Trend similarity | Average force under QS model | Average force under experiment | Error |
|---|---|---|---|---|---|
| Lift | 6.663% | 0.988 | 0.175 | 0.191 | 8.278% |
| Drag | 6.888% | 0.908 | 0.222 | 0.237 | 6.415% |

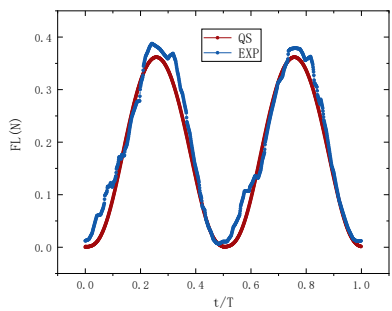
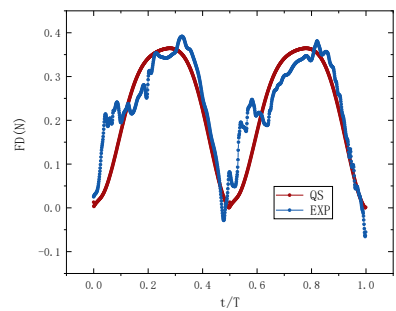

(a) Comparison results for lift  (b) Comparison results for drag

**Fig. 49.** Validation results

# Appendix C  Influence coefficients in tandem wing configurations

Drawing upon existing research[45] on aerodynamic interference in tandem wings, where the phase difference is 180 degrees, we extracted lift and drag data to analyze the interference effects on the forewings and hindwings. The analysis shown in the Fig. 50 revealed that:

1. Due to the phase difference of 180 degrees, the data exhibits a specific mirror effect between the forewings and hindwings.
2. The drag of the tandem wings influences both lift and thrust.

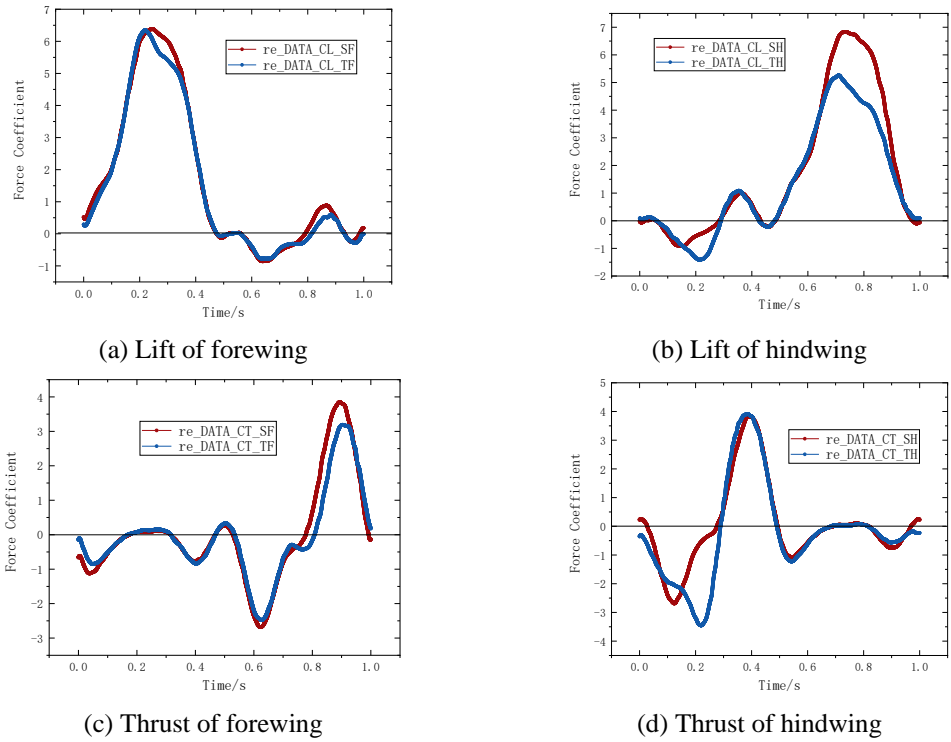

(a) Lift of forewing  (b) Lift of hindwing

(c) Thrust of forewing  (d) Thrust of hindwing

**Fig. 50.** Raw data for forces with and without tandem wing effects

To further analyze the impact of tandem wing interference on single-wing models, we established the following coefficients.

$$C_T = \frac{F_T - F_S}{F_S} \times 100\% \tag{102}$$

where $F_S$ is the single wing force, $F_T$ is the force under the single wing force under tandem wing configurations.

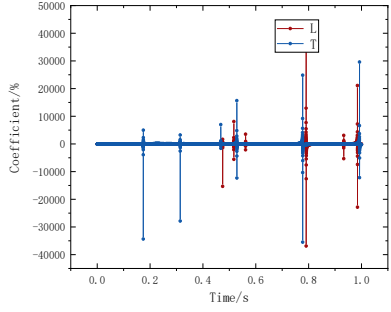 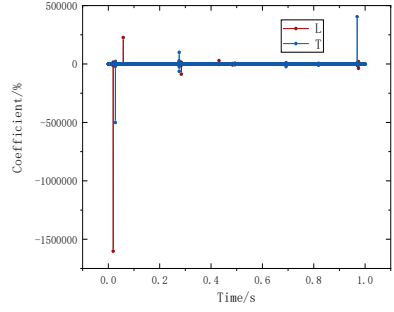

(a) The lift and thrust of the forewing    (b) The lift and thrust of the hindwing

**Fig. 51.** Date before treatment are jointly affected by the tandem wing interference.

Notably, as shown in Fig. 51, regions approaching zero exhibited sharp value increases. These abrupt changes, resulting from the 1/x relationship, signify rapid ratio increases. To mitigate these anomalies, we utilized the growth gradient as a criterion to exclude these regions. Considering that the aerodynamic forces in these areas are zero or near zero, discussing the ratio of tandem wing aerodynamic interference is irrelevant, thus setting these regions to zero. With a gradient threshold of 0.00001, we obtained data with the abrupt changes removed, as shown in Fig. 52.

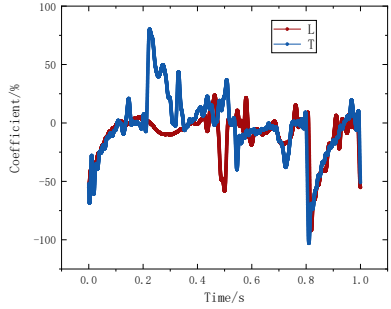 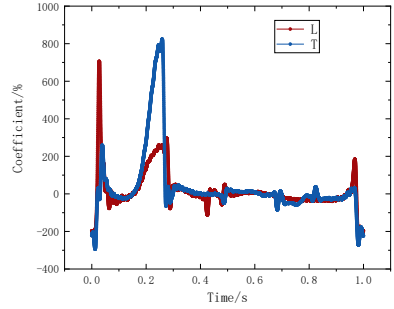

(a) The lift and thrust of the forewing    (b) The lift and thrust of the hindwing

**Fig. 52.** Date after treatments are jointly affected by the tandem wing interference.

Our key findings include:
1. Distinct variation patterns between the forewings and hindwings indicate the need for individualized models to capture the dynamics of tandem wing interference accurately.
2. Despite these differences, the influence of tandem wing interference on the lift and drag characteristics of both wings exhibited similar patterns. This observation led us to hypothesize that a unified model can approximate the effects of tandem wing interference on lift and drag.

Assuming the existence of a general tandem wing interference model, we propose the following formulation:

$$F_T = f(\phi_f, \dot{\phi}_f, \phi_h, \dot{\phi}_h) \cdot F_S \tag{103}$$

This model, inspired by existing aerodynamic frameworks, is further refined to:

$$F_T = f(X_0, X_1, X_2, X_3, X_4, X_5, X_6, X_7, X_8, X_9, X_{10}, X_{11}) \cdot F_S \tag{104}$$

$$X_0 = \dot{\phi}_f / (w_{maxf} \cdot C_{am}) \tag{105}$$

$$X_1 = \dot{\phi}_h / (w_{maxh} \cdot C_{am}) \tag{106}$$

$$X_2 = (\dot{\phi}_f - \dot{\phi}_h) / (w_{maxd} \cdot C_{am}) \tag{107}$$

$$X_3 = \phi_f / C_{am} \tag{108}$$

$$X_4 = \phi_h / C_{am} \tag{109}$$

$$X_5 = (\phi_f - \phi_h) / C_{am} \tag{110}$$

$$X_6 = \sin(2 \cdot \phi_f \cdot (\pi / C_{am})) \tag{111}$$

$$X_7 = \sin(2 \cdot \phi_h \cdot (\pi / C_{am})) \tag{112}$$

$$X_8 = \sin(4 \cdot \phi_f \cdot (\pi / C_{am})) \tag{113}$$

$$X_9 = \sin(4 \cdot \phi_h \cdot (\pi / C_{am})) \tag{114}$$

$$X_{10} = \sin(8 \cdot \phi_f \cdot (\pi / C_{am})) \tag{115}$$

$$X_{11} = \sin(8 \cdot \phi_h \cdot (\pi / C_{am})) \tag{116}$$

Utilizing data presented in Figure X, we employed symbolic regression to construct interference models for both fore and hindwings. The iterative process of symbolic regression demonstrated gradual convergence, as shown in Fig. 53, highlighting the effectiveness of this method in identifying underlying aerodynamic relationships.

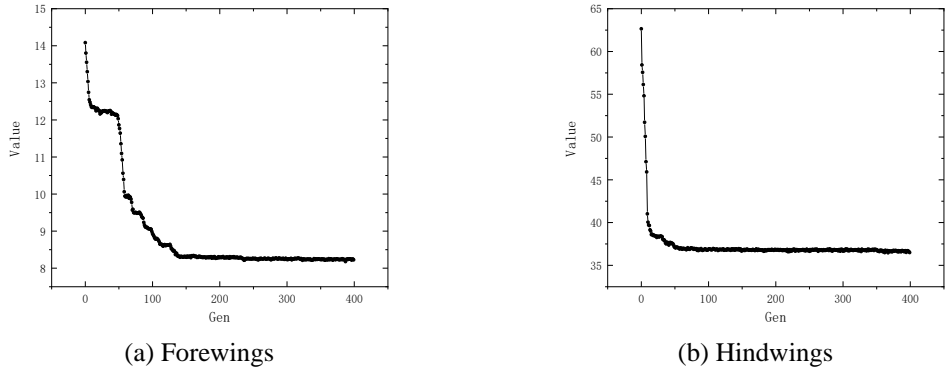

(a) Forewings  (b) Hindwings

**Fig. 53.** Iterative process of symbolic regression

The symbolic regression analysis yielded the following interference coefficients for the forewing:

$$\begin{aligned}
C_{TF} = & X_0 \cdot X_8 \cdot (-11.453 \cdot X_0 \cdot (-5 \cdot X_6 + X_7) - 22.906 \cdot X_1 \\
& + 11.453 \cdot X_2 - 11.453 \cdot X_4 + 11.453 \cdot X_5 - 11.453 \\
& \cdot X_7 - 11.453 \cdot \sin(\sin(X_8))) + X_0 \cdot X_9 \cdot (11.453 \\
& \cdot X_{10} + 11.453 \cdot X_{11} + 11.453 \cdot X_2 - 22.906 \cdot X_4 \\
& - 11.453 \cdot X_5 \cdot (X_1 - X_{11} - X_5) - 11.453 \cdot X_7) + 9 \\
& \cdot X_1 - 7.892 \cdot X_2 + 7.892 \cdot X_4 + X_8 - 6.166
\end{aligned} \tag{117}$$

The symbolic regression analysis yielded the following interference coefficients

for the hindwing:

$$\begin{aligned} C_{TH} = {} & 49.314 \cdot X_1 - (X_2 - X_5 - \sin(X_1 - 124.935)) \cdot (X_2 + X_4 - X_8 \\ & - 0.994) \cdot (X_3 + X_4 + X_6 + X_7) \cdot (-2X_1 - X_{10} - X_{11} \\ & + X_2 - X_7 \cdot (X_1 + X_7 - \sin(X_{10} - X_5)) \cdot (X_1 + X_{11} - X_9 \\ & - 45.822) - X_8 - (X_2 + X_4 - \sin(X_1 - 124.935))(2 \\ & \cdot X_0 - 3 \cdot X_{10} - X_9 - 34.855)(X_3 + X_4 + X_6 + X_7) \\ & - 49.314) \end{aligned} \quad (118)$$

A comparative analysis between the symbolic regression results and original data, as shown in Fig. 54, revealed a correlation coefficient of 0.88037 for the forewing and 0.88618 for the hindwing, relative to the average impact of interference. These coefficients validate the robustness of our proposed models in capturing the nuanced effects of tandem wing interference.

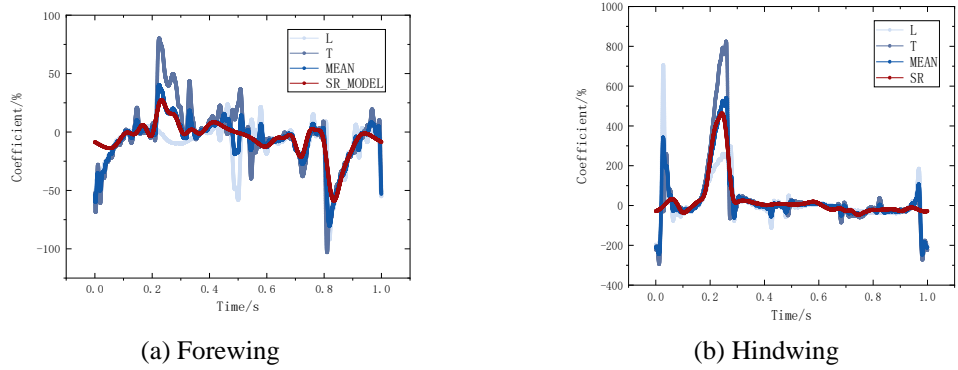

(a) Forewing　　　　　　　　　　(b) Hindwing

**Fig. 54.** Comparison of the construction results of the tandem wing effect on the wing based on symbolic regression